\documentclass[runningheads]{llncs}
\usepackage[T1]{fontenc}
\usepackage{graphicx}
\usepackage{amsmath,amsfonts}
\usepackage{algorithmic}
\usepackage{gensymb}
\usepackage{algorithm}
\usepackage{array}
\usepackage{subfig}
\usepackage{ulem}
\usepackage{textcomp}
\usepackage{stfloats}
\usepackage{xcolor}
\usepackage{url}
\usepackage{verbatim}
\usepackage{graphicx}
\usepackage{cite}
\usepackage{amssymb}
\usepackage{graphics}
\usepackage{caption}
\usepackage{multirow}
\usepackage{booktabs}
\usepackage{marvosym}

\begin{document}
\title{MVFAN: Multi-View Feature Assisted Network for 4D Radar Object Detection}
\titlerunning{MVFAN}
% % If the paper title is too long for the running head, you can set
% % an abbreviated paper title here
% %

%
\author{Qiao Yan\orcidID{0000-0003-0888-2047} \and
Yihan Wang\orcidID{0000-0002-9603-842X}\textsuperscript{\Letter}}
\authorrunning{Q. Yan and Y. Wang}
\institute{Nanyang Technological University, 639798, Singapore\\
\email{\{QIAO003,WANG1517\}@e.ntu.edu.sg}}
\maketitle              % typeset the header of the contribution
% %

\begin{abstract}
4D radar is recognized for its resilience and cost-effectiveness under adverse weather conditions, thus playing a pivotal role in autonomous driving. While cameras and LiDAR are typically the primary sensors used in perception modules for autonomous vehicles, radar serves as a valuable supplementary sensor. Unlike LiDAR and cameras, radar remains unimpaired by harsh weather conditions, thereby offering a dependable alternative in challenging environments. Developing radar-based 3D object detection not only augments the competency of autonomous vehicles but also provides economic benefits. In response, we propose the Multi-View Feature Assisted Network (\textit{MVFAN}), an end-to-end, anchor-free, and single-stage framework for 4D-radar-based 3D object detection for autonomous vehicles. We tackle the issue of insufficient feature utilization by introducing a novel Position Map Generation module to enhance feature learning by reweighing foreground and background points, and their features, considering the irregular distribution of radar point clouds. Additionally, we propose a pioneering backbone, the Radar Feature Assisted backbone, explicitly crafted to fully exploit the valuable Doppler velocity and reflectivity data provided by the 4D radar sensor. Comprehensive experiments and ablation studies carried out on Astyx and VoD datasets attest to the efficacy of our framework. The incorporation of Doppler velocity and RCS reflectivity dramatically improves the detection performance for small moving objects such as pedestrians and cyclists. Consequently, our approach culminates in a highly optimized 4D-radar-based 3D object detection capability for autonomous driving systems, setting a new standard in the field.

\end{abstract}

\section{Introduction}\label{sec_introduction}

\begin{figure}[t]
    \centering
    \includegraphics[width=0.5\linewidth]{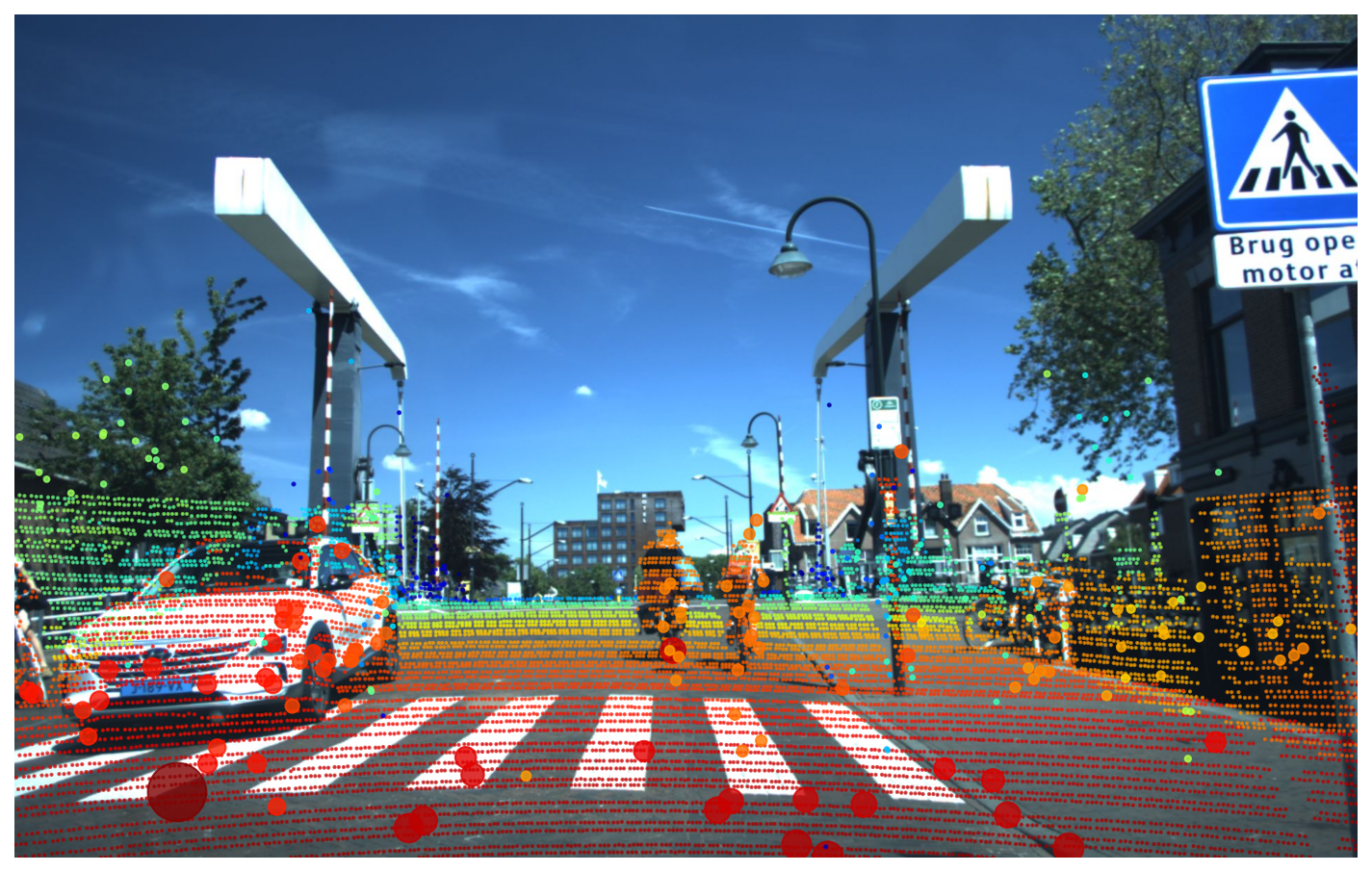}
    \caption{Visualization of one scene from VoD. The radar point clouds are platted as large dots while the LiDAR point clouds are shown in lines of small dots.}
    \label{fig_intro}
\end{figure}

A robust autonomous driving system consists of several modules, such as environment perception, path planning, decision-making, and control. Perception is the most critical module, as it informs the others. Traditionally, perception relies on the RGB camera, LiDAR, and radar, but camera and LiDAR struggle in adverse weather conditions. In contrast, radar is adaptable under all weather conditions and is indispensable for autonomous driving\cite{mmWaveRadar}. Conventional automotive radar provides limited information, restricting it to short-range collision detection. However, with the advent of millimeter-wave 4D radar, offering comprehensive $x, y, z$ position, and Doppler velocity information similar to LiDAR, this technology is considered ideal for perception applications. One example of radar point clouds is shown in Fig.\ref{fig_intro}.

Despite radar's broad use, its processed signals usually focus on position and speed, with a limited number of research on 4D-radar-based 3D object detection methods. \cite{CenterFusion, schumann2018semantic, 2DCar} explore semantic information for 2D object detection tasks. Some works\cite{CramNet, T-RODNet, Graph3D, 3DRadarCube, rodnet} utilize radial-frequency maps to identify potential 2D objects. Comprehensive benchmarks for 4D radar-based 3D object detection have emerged only recently with \cite{VOD, TJ4DRadSet,kradar}, while \cite{Astyx} is limited in size and scope. Prevailing strategies primarily employ LiDAR-based 3D detectors to 4D radar points. For example, methods like \cite{VOD, TJ4DRadSet} adapt existing techniques like PointPillars \cite{pointpillar}, and \cite{kradar} uses hierarchical 3D sparse and 2D convolution layers. 

However, these methods have yet to fully capitalize on the Doppler velocity and reflectivity of radar for 3D object detection, both of which are vital information captured by radar. While \cite{RadarNet} acknowledges the importance of the measured Doppler velocity and employs this data to align the radial velocity, it does not utilize this information for object detection. Furthermore, these studies commonly neglect the issue of the sparsity and irregularity of radar point clouds compared to LiDAR. We observe a marked performance discrepancy when implementing PointPillars with two different data inputs within the same dataset, VoD\cite{VOD}. This inconsistency infers that the direct implementation of LiDAR-based detectors on radar points is inappropriate. 

Given that Doppler velocity and reflectivity are two essential types of information overlooked by existing methods, we propose a novel backbone named the Radar Feature Assisted Backbone, designed to fully integrate these data into our framework. The incorporation of Doppler velocity and reflectivity data enables our network to more effectively discern potential objects. This stems from the fact that points on moving objects often display distinct Doppler velocities and reflectivity, setting them apart from points in the surrounding environment. By leveraging these distinctive characteristics, our proposed backbone significantly improves the network's capacity to precisely differentiate and identify objects.

To compensate for the sparsity of radar point clouds, we propose a multi-view-based framework. We construct both BEV and cylinder pillars to extract features hierarchically. In comparison to single-modal methods, such as point-based and grid-based approaches, multi-view strategies yield richer feature maps, which are crucial for handling the sparse characteristics of 4D radar point clouds. To address the irregularity issue, we introduce a novel component called the Positional Map, which assigns different weights to foreground and background points. This ensures that foreground points, which are more likely to be reflections from detected objects, are given higher importance, while background points receive lower weights. This approach enables us to prioritize detecting objects measured by foreground points, compensating for the irregularity of radar point clouds.

\textbf{Contributions.} Overall, our contributions can be summarized as follows:

\begin{itemize}
\item We propose the Multi-View Feature Assisted Network (\textit{MVFAN}), an end-to-end, anchor-free, and single-stage framework for sufficient feature utilization.
\item We introduce a novel Position Map Generation module that enhances feature learning by reweighing foreground and background points and their features, addressing the irregular distribution of radar point clouds.
\item We propose a novel backbone, named Radar Feature Assisted backbone, that fully leverages the significant information provided by the Doppler velocity and RCS reflectivity of the 4D radar sensor.

% , which has not been fully utilized in existing radar-based 3D object detection frameworks.
\item We conduct comprehensive experiments on Astyx\cite{Astyx} and VoD\cite{VOD} datasets with ablation studies to demonstrate the effectiveness of our proposed modules.
\end{itemize}

\section{Related Works}\label{sec_LR}
% \subsection{Radar-based Object Detection}\label{sec_OD_on_radar}
{\bf{2D object detection on radar.}} Currently, there are few works focusing on end-to-end 3D object detection on radar point clouds. Some works aim to regress 2D bounding boxes without elevation and height of detected objects. In \cite{2DCar}, PointNet \cite{pointnet} is applied to extract radar features and generate 2D bounding box proposals together to extract features for 2D object detection. Similarly, \cite{schumann2018semantic} performs semantic segmentation on radar points and then clustered points are utilized to segment objects to regress 2D bounding boxes. Some works utilize radar points as complementary information and fuse with other sensors. Nabati et al.\cite{RRPN} proposes a two-stage region proposal network. To generate anchor or region of interest, radar points are projected into camera coordinates and distance of radar points are used to properly crop the image as a learned parameter. Later, they \cite{CenterFusion} further fine-tune this model to operate on point cloud and image. Radar points are expanded as pillars and associated with extracted image features by frustum view. However, because of the lack of height information caused by pillar expansion, 3D bounding boxes are only estimated and the evaluation results are far from satisfactory, only around 0.524\% mAP (mean average precision) on nuSences dataset\cite{nuScenes}. In \cite{RRPN, CenterFusion}, radar point clouds are only used as auxiliary input. \cite{SeeThroughFog} proposes an adaptive approach that deeply fuses information from radar, LiDAR and camera, which is proven to be efficient under foggy environments. \cite{AM3Net} proposes possibility for data fusion of radar images. Also since this dataset employs 3D radar with low spatial resolution, only 2D objects are regressed in image plane. Given that radar and LiDAR obtain point clouds in 3D world coordinate, some mechanisms to exploit radar and LiDAR for perception have merged. RadarNet \cite{RadarNet} and MVDNet \cite{MVD} both fuse sparse radar point clouds and dense LiDAR point clouds after feature extraction at an early stage by voxelization of point cloud in BEV coordinates. They both achieve optimal performance in regressing 2D object detection under BEV frames in nuScence, DENSE\cite{SeeThroughFog} and ORR\cite{RadarRobotCarDatasetICRA2020}.

{\bf{3D object detection on radar.}} Few works exploit radar point clouds for 3D object detection. In \cite{RoadUserDetection}, RTCnet is first embedded to encode the range-azimuth-Doppler image to output multi-class features and then it performs clustering technique to gather object points. Some \cite{Astyx, TJ4DRadSet, VOD} only implement LiDAR-based detectors such as PointPillars \cite{pointpillar} for 3D detection with raw radar point clouds as input. \cite{3D_Radar_Camera} considers the sparsity of radar point cloud and exploits previously designed method \cite{Joint_3D} with radar and camera as input for 3D detection. This proposal level fusion method highly depends on the accuracy of each feature extractor of radar and camera. Xu et al. \cite{RPFA-Net} carefully designed a self-attention technique to extract radar features with PointPillars as the backbone and baseline for evaluation while only considering global features. The major drawbacks of all these methodologies are that they do not carefully take the sparsity and ambiguity of measured radar points into account and do not fully make use of Doppler and RCS measured by radar.

\section{Methodology}\label{sec_method}

\subsection{Overview}\label{sec_overview}

\begin{figure*}[t]
    \centering
    \includegraphics[width=1\linewidth]{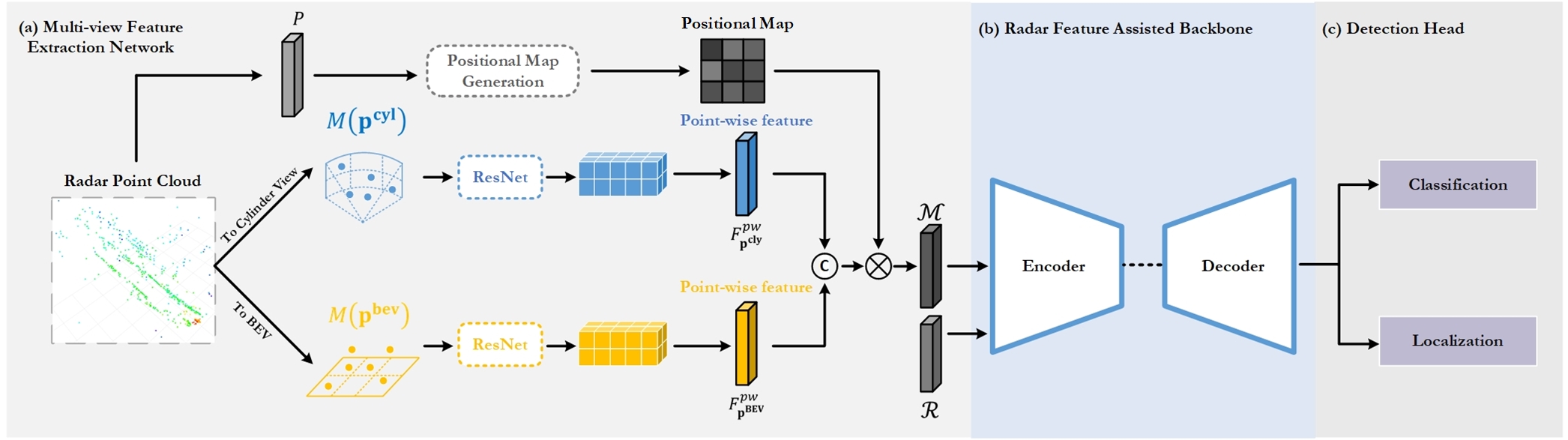}
    \caption{The flowchart of \textit{MVFAN}.}
    \label{fig_flowchart}
\end{figure*}

As shown in Fig.\ref{fig_flowchart}, \textit{MVFAN} consists of three continuously connected modules: (a) Multi-View Feature Extraction Network. The raw radar point cloud is firstly transformed into cylindrical and BEV pillar as pseudo images, $M(\mathbf{p^{cyl}})$ and $M(\mathbf{p^{BEV}})$. Pillars are then encoded by ResNet blocks to extract high dimensional features and then projected back to points. Multi-view point-wise features are embedded to construct a fused feature map $\mathcal{M}$ after reweighing by Positional Map. In parallel, a radar feature vector $\mathcal{R}$ is built. (b) Radar Feature Assisted Backbone. In the backbone, $\mathcal{M}$ is aided by $\mathcal{R}$ to encode radar feature map assisted by convolution with radar Doppler and RCS representations to build dense map $\mathcal{DM}$ in a top-down and upsampling backbone. (c) Detection Head. Finally, 3D bounding boxes are regressed through the detection head by $\mathcal{DM}$.

% (a) The raw radar point cloud is firstly transformed into cylindrical and BEV pillar as pseudo images, $M(\mathbf{p^{cyl}})$ and $M(\mathbf{p^{BEV}})$. Pillars are then encoded by ResNet blocks to extract high dimensional features and then projected back to points. Multi-view point-wise features are embedded to construct a fused feature map $\mathcal{M}$ after reweighing by Positional Map. In parallel, a radar feature vector $\mathcal{R}$ is built. (b) In radar feature assisted backbone, $\mathcal{M}$ is aided by $\mathcal{R}$ to encode radar feature map assisted by convolution with radar Doppler and RCS representations to built dense map $\mathcal{DM}$ in a top-down and upsampling backbone. (c) 3D bounding boxes are regressed through detection head by $\mathcal{DM}$.

\textbf{(a) Multi-View Feature Extraction Network.} Given a single frame of a raw point cloud, firstly points are pre-processed by transforming into cylinder and BEV grid cells then stacked into pillars as BEV pseudo image synchronously. Then, two individual feature extractors, ResNet blocks \cite{resnet}, are engaged to extract high-dimensional and complementary representative features to boost auxiliary point-wise and pillar-wise features. In order to stack and augment the feature map in a point-wise manner, previously extracted pillar features are transformed back to the point view by established pillar-to-point and point-to-pillar mapping along with the position information of raw points. Therefore, the features of each individual point are aggregated by concatenating abstracted semantic features which are generated from both the cylinder view and the bird's-eye view. Next, we generate Positional Map to reweigh the features. This process allows for a more distinct separation of foreground points from background points.

\textbf{(b) Radar Feature Assisted Backbone.} The reweighted feature map together with the Doppler velocity and reflectivity are fed into the Radar Feature Assisted Backbone with an encoder-decoder structure, which heuristically learns the contextual representative feature and geometric characteristics with positional encoding. The radar-assisted feature guides the learning process by summarizing the point-wise features through measured Doppler velocity and reflectivity of each point. The downsampling and upsampling modules of the backbone enable pyramid feature learning and facilitate the extraction of features at multiple scales and receptive fields.

% The effect of raw information, as well as Doppler velocity and RCS reflectivity, have not been comprehensively explored by any of the existing 3D detectors. We define this information as radar-assisted feature $\mathcal{R}$ and propose a Radar Feature Assisted Backbone to make full of it. The reweighted feature $\mathcal{M}$ in the higher level and the radar-assisted feature $\mathcal{R}$ in the lower level are embedded into our backbone with an encoder-decoder structure, which heuristically learns the contextual representative feature and geometric characteristics with positional encoding. The radar-assisted feature $\mathcal{R}$ guides the learning process by summarizing the point-wise features through positional encoding with their 3D information. The downsampling and upsampling modules of the backbone enable pyramid feature learning and facilitate the extraction of features at multiple scales and receptive fields.

\textbf{(c) Detection Head. } Eventually, the multiscale features obtained from the backbone are fed into the detection head where 3D positions, height, weight, length, and heading angles of each box are predicted precisely.

% These representative points are then aggregated and projected into bird's-eye view again to construct a fused feature map. Such dense feature map does not only accommodate geometric information but also holds abundant learned semantic features. Next, in our backbone, this concatenated feature map is forwarded into 2D CNN backbone together with previously extracted layers of radar feature map in proposed assisted-convolution block. In the meanwhile, integrated feature map passes through our top-down networks aiming at creating context sub-feature map with multiple scales at different stages.  Three layers of radar feature map are maxpooled sequentially to ensure constant spatial resolution.  Each layers of maxpooled features and sub-feature map at each stage is combined together to ensure the point-wise and pillar-wise correspondences keep consistent. The high-level features are finally assembled by upsampling and concatenating the multi-scale top-down features before embedded to regress precise classification, localization of detected objects.

\subsection{Multi-View Feature Extraction}\label{sec_multiview}

\subsubsection{BEV Pillar Projection}\label{sec_pillar_projection}

% Similarly to PointPillars\cite{pointpillar} and PillarOD \cite{Pillar-based}, the pillar-to-point and point-to-pillar mapping should be established as preliminaries. Given raw points $P=\{{p_n}\}^{N-1}_{n=0}$ where $p_n=(x_n, y_n, z_n)$ in Cartesian coordinate and the detection range $[X_{min}, X_{max}], [Y_{min}, Y_{max}], [Z_{min}, Z_{max}]$ with spatial resolution in BEV, $R^x, R^y$. These points are primarily transformed into bird's-eye view and then stacked as pillars in grid cells along $Z$ axis. The grid cell indices of point $p_n$ are expressed as Eq.\eqref{point-to-pillar} where point $p_n$ is assigned to pillar $\mathbf{p}_m$.

% Similarly to PointPillars\cite{pointpillar} and PillarOD \cite{Pillar-based}, the pillar-to-point and point-to-pillar mapping should be established as preliminaries. Given raw points $P=\{{p_n}\}^{N-1}_{n=0}$ where $p_n=(x_n, y_n, z_n)$ in Cartesian coordinate, every point $p_n$ is assigned to the corresponding pillar $\mathbf{p}_m$ according to the point-to-pillar mapping as Eq.\eqref{point-to-pillar}.
Similar to PointPillars \cite{pointpillar} and PillarOD \cite{Pillar-based}. Given a set of raw points $P=\{{p_n}\}^{N-1}_{n=0}$ in Cartesian coordinate, where $p_n=(x_n, y_n, z_n)$, each point $p_n$ is assigned to the corresponding pillar $\mathbf{p}_m$ based on the point-to-pillar mapping described by Eq.\eqref{point-to-pillar}.

\begin{equation}
\begin{split}
\label{point-to-pillar}
F_{\mathbf{p}^{BEV}}(p_n^{BEV}) = \mathbf{p}_m^{BEV} 
\end{split}
\end{equation}

However, the mapping from pillar to points can be expressed as in Eq.\eqref{pillar-to-point}, where each pillar encapsulates all points within its boundaries.

\begin{equation}
\label{pillar-to-point}
F_{p^{BEV}}(\mathbf{p}_m^{BEV}) = \{p_n^{BEV} \, | \, \forall p_n^{BEV} \,\in \,  \mathbf{p}_m^{BEV}\}
\end{equation}
Here, we define $F_\mathbf{p}(p_n)$ as the point-to-pillar mapping which returns the pillar $\mathbf{p}_m$ that contains point $p_n$, and $F_p(\mathbf{p}_m)$ as the pillar-to-point mapping which determines the set of points within pillar $\mathbf{p}_m$ by using $F_\mathbf{p}$. Thus, the point-to-pillar and pillar-to-point mapping are established, respectively, as described in Eq.\eqref{point-to-pillar} and Eq.\eqref{pillar-to-point}.

% Here, we define $F_\mathbf{p}(p_n)$ as the point-to-pillar mapping which returns the pillar $\mathbf{p}_m$ of point $p_n$ and $F_p(\mathbf{p}_m)$ as the pillar-to-point mapping which determines the corresponding pillar $\mathbf{p}_m$  by $F_\mathbf{p}$ and converges a set of points within this pillar $\mathbf{p}_m$. Thus, point-to-pillar and pillar-to-point mapping is established respectively as Eq.\eqref{point-to-pillar} and Eq.\eqref{pillar-to-point}.
\subsubsection{Cylinder Pillar Projection}\label{sec_cylinder_project} \cite{MVF} discovered that the spherical projection of point clouds into a range image can result in dispensable distortion in the $Z$-axis, i.e., the height information in Cartesian coordinate, and this can negatively impact the performance as pointed out by previous works\cite{Pillar-based, hollow_3D}. To prevent information loss, we use cylinder projection of raw point clouds as complementary representations without perspective distortion. The formulation for cylinder projection is given by Eq.\eqref{cylinder}.
% It has been discovered that in MVF\cite{MVF} the spherical projection of point clouds into range image can lead to dispensable distortion in $Z$-axis, the height information in Cartesian coordinate and will avoidably devalue the performance as pointed out by \cite{Pillar-based, hollow_3D}. Accordingly, we conduct cylinder projection of raw point clouds as complementary representations without perspective distortion with the following formulation Eq.\eqref{cylinder} to prevent information loss.

\begin{equation}
\label{cylinder}
\rho_n = \sqrt{x_n^2 + y_n^2},\quad \phi_n = arctan(\frac{y_n}{x_n}),\quad z_n^{\prime} = z_n
\end{equation}

% Given a raw point cloud $p_n$ in Cartesian coordinate $(x_n, y_n, z_n)$, the corresponding coordinate under cylinder coordinate can be described as $p_n^{cyl} = (\rho_n, \phi_n, z_n^{\prime})$, where $\rho$ and $\phi$ are the distance, the azimuth angle of the point cloud and $z^{\prime}$ retains the height particularly. By setting the partition range under cylinder coordinate, points under Cartesian coordinates are first transformed into cylinder coordinate and then projected as cylinder images. With the same detection range under Cartesian coordinate as $[X_{min}, X_{max}], [Y_{min}, Y_{max}], [Z_{min}, Z_{max}]$, the range in cylinder coordinate is fixed at $[P_{min}, P_{max}], [\Phi_{min}, \Phi_{max}], [Z^{\prime}_{min}, Z^{\prime}_{max}]$ by Eq.\eqref{cylinder}. Similarly as in Section \ref{sec_pillar_projection} with fixed spatial resolution $R^{\phi}, R^{z^{\prime}}$, the cylinder view points are reconstructed into pillar and pillar-to-point mapping is organized as Eq.\eqref{point_to_pillar_cylinder} and Eq.\eqref{pillar_to_point_cylinder}.

Given a raw point cloud $p_n$ in Cartesian coordinate $(x_n, y_n, z_n)$, its corresponding coordinate under cylinder coordinate can be described as $p_n^{cyl} = (\rho_n, \phi_n, z_n^{\prime})$, where $\rho$ and $\phi$ represent the distance and azimuth angle of the point cloud, respectively, and $z^{\prime}$ retains the height information. As in Section \ref{sec_pillar_projection}, all points under the cylinder view are reconstructed into pillars, and pillar-to-point mapping is established according to Eq.\eqref{point_to_pillar_cylinder} and Eq.\eqref{pillar_to_point_cylinder}.

% Given a raw point cloud $p_n$ in Cartesian coordinate $(x_n, y_n, z_n)$, the corresponding coordinate under cylinder coordinate can be described as $p_n^{cyl} = (\rho_n, \phi_n, z_n^{\prime})$, where $\rho$ and $\phi$ are the distance, the azimuth angle of the point cloud and $z^{\prime}$ retains the height particularly. Similarly as in Section \ref{sec_pillar_projection}, all points under cylinder view are reconstructed into pillars, and pillar-to-point mapping is organized as Eq.\eqref{point_to_pillar_cylinder} and Eq.\eqref{pillar_to_point_cylinder}.

\begin{equation}
\begin{split}
\label{point_to_pillar_cylinder}
F_{\mathbf{p}^{cyl}}(p_n^{cyl}) = \mathbf{p}^{cyl}_m 
\end{split}
\end{equation}

\begin{equation}
\label{pillar_to_point_cylinder}
F_{p^{cyl}}(\mathbf{p}_m^{cyl}) = \{p_n^{cyl} \, | \, \forall p_n^{cyl} \,\in \,  \mathbf{p}_m^{cyl}\}
\end{equation}

\begin{figure}[t]
    \centering
    \subfloat[BEV Pillar Projection\label{fig_BEV_project}]{%
        \includegraphics[width=0.4\linewidth]{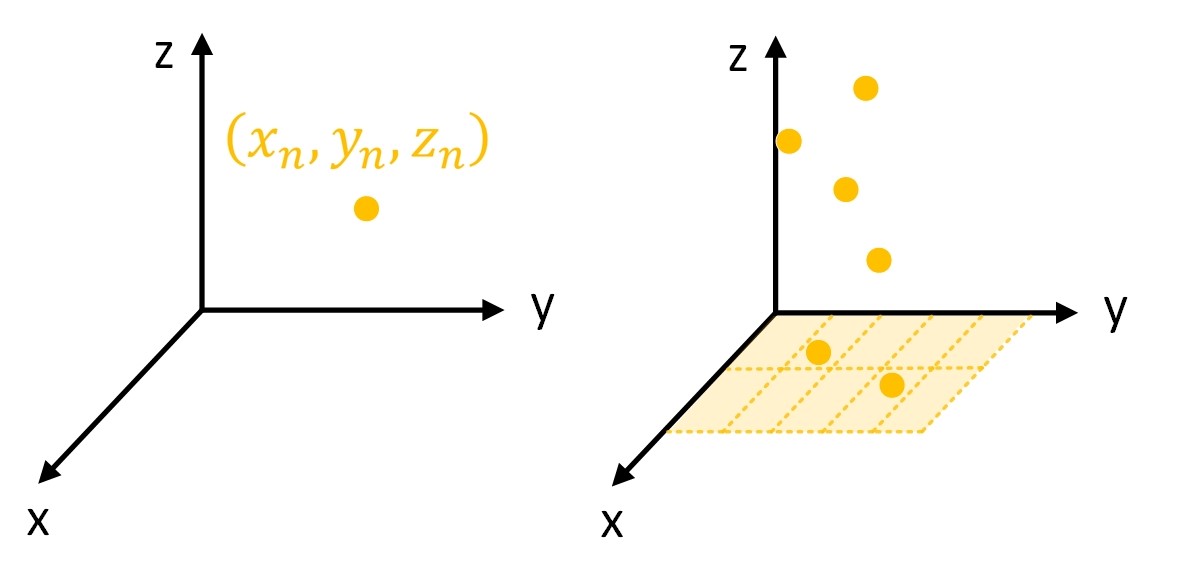}}
    \hfill
    \subfloat[Cylinder Pillar Projection\label{fig_cylinder_projection}]{%
        \includegraphics[width=0.4\linewidth]{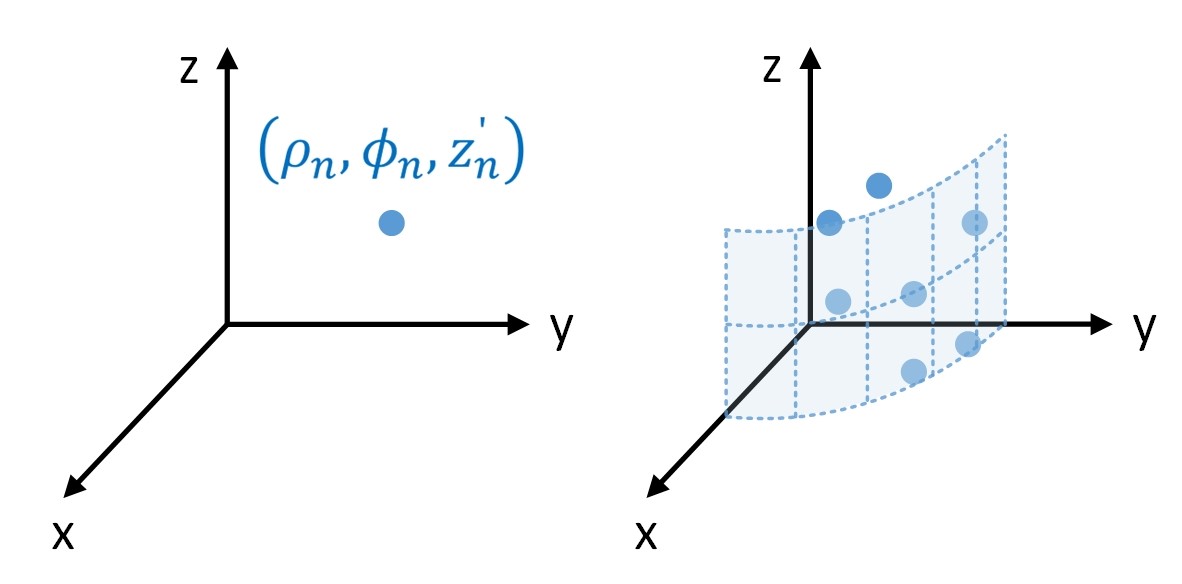}}
\end{figure}

\subsubsection{Multi-View Feature Extraction}\label{sec_multiview_feature}

% After points $P=\{{p_n}\}^{N-1}_{n=0}$ converted into BEV and cylindrical pillars $\mathbf{p^{BEV}}$ and $\mathbf{p^{cyl}}$, these pillars are stacked accordingly as BEV and cylindrical pseudo images $M(\mathbf{p^{BEV}})$ and $M(\mathbf{p^{cyl}})$. For the reason that $M(\mathbf{p^{BEV}})$ and $M(\mathbf{p^{cyl}})$ can be regarded as images, we can natively believe they can also be encoded by 2D CNN. Consequently, we exercise ResNet \cite{resnet} to extract semantic high-dimensional features. We denote these pillar-wise features as $ f_{\mathbf{p}_m^{BEV}}$ and $ f_{\mathbf{p}_m^{cyl}}$ for BEV and cylinder pillar $\mathbf{p}_m^{BEV}$ and $\mathbf{p}_m^{cyl}$.

After converting the points $P=\{{p_n}\}^{N-1}_{n=0}$ into BEV and cylindrical pillars $\mathbf{p}^{BEV}$ and $\mathbf{p}^{cyl}$, we stack them to form BEV and cylindrical pseudo images $M(\mathbf{p}^{BEV})$ and $M(\mathbf{p}^{cyl})$. Since $M(\mathbf{p}^{BEV})$ and $M(\mathbf{p}^{cyl})$ can be treated as images, we can use 2D CNNs to extract high-dimensional features. Therefore, we employ ResNet \cite{resnet} to extract semantic features on a per-pillar basis, which we denote as $f_{\mathbf{p}_m^{BEV}}$ and $ f_{\mathbf{p}_m^{cyl}}$ for the BEV and cylindrical pillars $\mathbf{p}_m^{BEV}$ and $\mathbf{p}_m^{cyl}$, respectively.

\begin{equation}
\label{resnet_features}
\begin{gathered}
 f_{\mathbf{p}_m^{BEV}}=ResNet(M(\mathbf{p}_m^{BEV})) \\ f_{\mathbf{p}_m^{cyl}}=ResNet(M(\mathbf{p}_m^{cyl}))   
\end{gathered}
\end{equation}

These high-level features are projected back as point-wise features with corresponding points according to the constructed pillar-to-point relation. To restore point-wise features from pillars, the pillar-to-point feature mapping is established as Eq.\eqref{pillar_to_point_feature} where $f_{p_n^{BEV}}$ and $f_{p_n^{cyl}}$ denote the point-wise features.

\begin{equation}
\label{pillar_to_point_feature}
\begin{gathered}
f_{p_n^{BEV}} = f_{\mathbf{p}_m^{BEV}},\text{where}\quad \mathbf{p}_m^{BEV} =F_{\mathbf{p}^{BEV}}(p_n^{BEV})   \\
f_{p_n^{cyl}} = f_{\mathbf{p}_m^{cyl}},\text{where}\quad \mathbf{p}_m^{cyl} =F_{\mathbf{p}^{cyl}}(p_n^{cyl})   
\end{gathered}
\end{equation}

Finally, $d$ dimensional points-wise features for $N$ number of points are gathered from both the BEV and cylinder view as expressed in Eq.\eqref{point-wise-N-d}.

\begin{equation}
\label{point-wise-N-d}
\begin{gathered}
    F^{pw}_{p^{BEV}} =  \{{f_{p_n^{BEV}}}\}^{N-1}_{n=0}\subseteq \mathbb{R}^{N\times d} \\
    F^{pw}_{p^{cyl}} =  \{{f_{p_n^{cyl}}}\}^{N-1}_{n=0}\subseteq \mathbb{R}^{N\times d}
\end{gathered}
\end{equation}

% Therefore $d$ dimensional $N$ number of points-wise features, $f_{p_n^{BEV}}\subseteq \mathbb{R}^{N\times d}$ and $F^{pw}_{\mathbf{p}^{cyl}}(p_n^{cyl})\subseteq \mathbb{R}^{N\times d}$ are gathered by Eq.\eqref{point-wise-gather-BEV} and Eq.\eqref{point-wise-gather-cylinder}.

% \begin{equation}
% \label{point-wise-gather-BEV}
% f_{p_n}^{BEV} = ResNet(M(\mathbf{p}_m^{BEV}))
% \end{equation}

% \begin{equation}
% \label{point-wise-gather-cylinder}
% F^{pw}_{\mathbf{p}^{cyl}}(p_n^{cyl}) = ResNet(M(\mathbf{p}_m^{cyl}))
% \end{equation}

% While point-wise features are scattered in Cartesian coordinate as points with $C$ dimensional features, they are accumulated once more and projected into BEV as Section \ref{sec_pillar_projection} to create a fused feature map $\mathcal{M}$ with $(C, H, W)$ where $C$ is the dimension of the output feature by Eq.\eqref{resnet_features} and $(H, W)$ is the feature map size by point-to-pillar map as Eq.\eqref{point-to-pillar}. Concurrently a pillar feature map  $\mathcal{R}$ with shape $(3, H, W)$ is also piled up by BEV pillar project where $3$ contains the Doppler velocity $v$, Doppler velocity compensated by ego motion $v_r$ and RCS reflectivity generated by the working mechanism of radar, as radar feature, with the same horizontal shape as feature map $\mathcal{M}$ as Eq.\eqref{M_R}.

% \begin{equation}
% \label{M_R}
% \mathcal{R} = M(\mathbf{p}^{BEV}))[v, v_r, RCS]
% \end{equation}

\subsubsection{Positional Map Generation}\label{sec_Positional_Map_Generation} To address the irregular distribution of radar point clouds, which is often overlooked by existing methods, we introduce the Positional Map Generation module that enables us to prioritize detecting objects captured by foreground points by assigning varying weights to the foreground and background points. In comparison to previous work RPFA-Net \cite{RPFA-Net} which only leverages pillar features aggregated by sparse points as insufficient feature representation, the Positional Map enhances feature characterization in both global and local scales. 

The point-wise features extracted from multi-view sources, $F^{pw}_{p^{BEV}}$ and $F^{pw}_{p^{cyl}}$, are then multiplied by the generated Positional Map. The resulting reweighted feature map is subsequently fed into the backbone for further processing. The point-wise classification result vector $A$ is regressed by the Auxiliary Loss module.

\begin{figure}[htbp]
    \centering
    \includegraphics[width=0.6\linewidth]{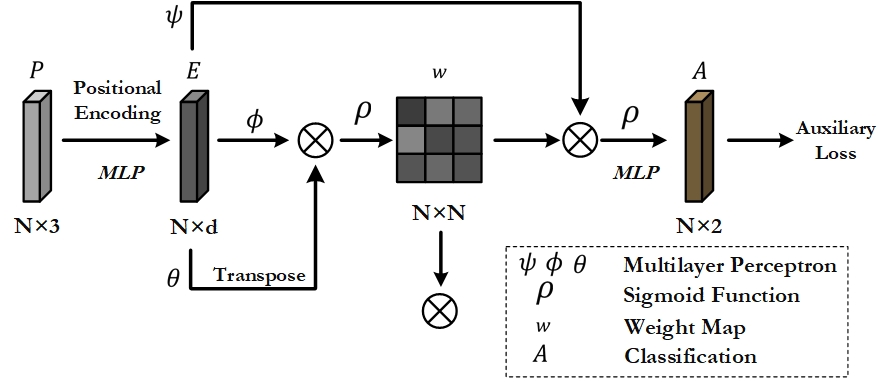}
    \caption{Positional Map Generation.}
    \label{fig_map_generation}
\end{figure}

\textbf{Positional Encoding. } To encode the positional information of the raw point clouds $P=\{{p_n}\}^{N-1}_{n=0}$, where $p_n=(x_n, y_n, z_n)$ represents the geometric position of the $n$-th point, we employ the K-nearest neighbors (KNN) algorithm to identify each center point $p_n^k$. We then compute the relative position of each point with respect to its KNN center as $||p_n-p_n^k||$, which is concatenated with the absolute position of each center point for positional encoding as in Eq. \eqref{positional_encoding}.

\begin{equation}
\label{positional_encoding}
e_n = MLP(p_n^k, ||p_n-p_n^k||),\, E=\{{e_n}\}^{N-1}_{n=0}
\end{equation}
$MLP$ is a multilayer perception that outputs a vector of dimension $d$. The resulting positional encoded vector $E\subseteq \mathbb{R}^{N\times d}$ captures both local and global geometric context, with the relative position $||p_n-p_n^k||$ encoding local information and the KNN center $p_n^k$ representing the global distribution of all radar points. The effectiveness of this positional encoding strategy in fusing local spatial information as essential feature representation has been demonstrated in \cite{3DCTN}. 
% With the relative position $||p_n-p_n^k||$ presenting the local information and the KNN center $p_n^k$ showing the global distribution of all radar points, the positional encoded vector $E\subseteq \mathbb{R}^{N\times d}$ is aggregated as local and global geometric context where $d$ is the output dimension of the multilayer perception ($MLP$). The positional encoding strategy is proven to be effective to fuse the local spatial information as essential feature representation in \cite{3DCTN}. Considering the sparsity of radar points, we add KNN center coordinates as global spatial features to enrich feature representations.

\textbf{Map Generation. }Next, we perform a matrix operation to generate a Positional Map from the positional encoded feature vector $E$, as illustrated in Fig.\ref{fig_map_generation}. The map $w$ is expressed as Eq.\eqref{map_generation} where $\rho$ is the sigmoid activation function and $\phi,\theta$ are point-wise transformations, two independent $MLPs$. $w\subseteq \mathbb{R}^{N\times N}$ is generated by calculating the correlation between the $i^{th}$ and $j^{th}$ positional encoded features of $E$, which captures the relationship between different points $p_i \in P$ and $p_j \in P$. The Positional Map $w$ is then multiplied by the concatenation of point-wise features $F^{pw}_{\mathbf{p}^{BEV}}(p_n^{BEV})$ and $F^{pw}_{\mathbf{p}^{cly}}(p_n^{cly})$, respectively, to re-weight these point-wise features according to the learned geometric information.

\begin{equation}\begin{gathered}
\label{map_generation}
w = \Sigma \rho(\phi(e_i)\theta(e_j)^T),\quad w\subseteq \mathbb{R}^{N\times N} \\
\mathcal{M}_0 = w\times Concat(F^{pw}_{p^{cly}}, F^{pw}_{p^{BEV}})
\end{gathered}
\end{equation}

The purpose of generating the Positional Map from the positional encoded feature vector is to capture the interdependence of local point clouds and learn high-dimensional features through the forward modules. This allows for the aggregation of both the geometric and feature context of the points, which in turn implies their relations in 3D position. By multiplication of concatenated point-wise features from multi-view, we obtain the geometric-related feature $\mathcal{M}$ which is fed into the backbone for further processing.

\textbf{Auxiliary Loss.} We introduce an auxiliary loss to help the learning process of the Positional Map by differentiating foreground and background points. After obtaining $w$, we use a $MLP$ to predict the point-wise classification result vector $A$ by performing matrix multiplication as Eq.\eqref{ax_loss}. The sigmoid function $\rho$ performs binary classification of foreground and background points. The $MLP$ layer $\psi$ takes the positional encoded feature vector $E$ as input. $L_i$ is the focal loss of point $p_i$ derived from the ground truth labels indicating whether $p_i$ is inside the 3D bounding boxes. The aim of the auxiliary loss is to give more importance to the foreground points that indicate possible objects, while the background points are assigned fewer proportions. The final auxiliary loss is obtained by taking the mean of all individual losses, as expressed in Eq.\eqref{ax_loss}.
% \textbf{Auxiliary Loss. }Intuitively, the foreground points are supposed to be given more propositions as they imply the possible objects whereas the background points are assigned with fewer proportions. To this end and to help the learning processing of the weight map, we introduce auxiliary loss to identify foreground and background points. After getting the weight map $w$, we perform matrix multiplication with a $MLP$ to predict the point-wise classification results vector $P$.
\begin{equation}\begin{gathered}
\label{ax_loss}
A=\rho(w\times \psi(E)),\quad A=\{a_{i,0},a_{i,1}\}_{i=0}^{N-1}, \quad A\subseteq \mathbb{R}^{N\times 2} \\
L_{ax} = \frac{1}{N} \Sigma_{i=0}^{N} L_i \\
L_i =
\begin{cases}
-\alpha_i (1-a_{i,1})^\gamma \log(a_{i,1}) & \text{if } y_i=1 \\
-(1-\alpha_i) a_{i,0}^\gamma \log(1-a_{i,0}) & \text{otherwise}
\end{cases}
\end{gathered}
\end{equation}
where $y_i$ is the ground truth label for point $p_i$ (1 for foreground, 0 for background), $a_{i,1}$ and $a_{i,0}$ are the predicted probabilities for foreground and background, respectively. $\alpha_i=0.25$ is a weight factor that assigns more importance to the foreground points, and $\gamma=2$ is a modulating factor that down-weights the loss for well-classified samples in focal loss \cite{focalloss}.

\textbf{Discussion. }The adoption of a multi-view strategy enhances feature learning and addresses the sparsity of radar point clouds by providing more comprehensive representations. Considering the irregular distribution of radar point clouds, we introduce a Positional Map, which predicts the binary classification of background and foreground points using a supervised auxiliary loss function. By applying the Positional Map, we emphasize the foreground points and their associated semantic features. This integration of geometric information in the Positional Map enhances performance and is particularly beneficial for handling the irregularity issue.

% where $\alpha$ is the $MLP$ layer and $\rho$ is the sigmoid function to perform binary classification of foreground and background points. $L_i$ is the focal loss of point $p_i$ derived from ground truth labels whether $p_i$ is inside the 3D bounding boxes.

% 3D medical: Although self-attention is able to model long-range dependencies over the global domain, it cannot aggregate local information
% Point Transformer: position encoding is important for both the attention generation branch and the feature transformation branch
% LFT-NET: The purpose of local feature transformer model is to capture the dependence of local point clouds and learn high-dimensional features through forward neural network
% the local feature transform model performs an embedding operation on the coordinates of the input point cloud through a local position encoding unit.
% positional encoding:fuse the local spatial information, which is essential for local feature representation. 
% PU-Transformer: Point Cloud Upsampling Transformer
% local information is aggregated from both the geometric and feature context of the points, implying their 3D positional relations
% In addition to the neighbors’ relative positions showing each point’s local detail, we also append the centroids’ positions in 3D space, indicating the global distribution for all points

\subsection{Radar Feature Assisted Backbone}\label{sec_backbone}

% LFT: can heuristically learn the geometric characteristics of local point clouds
% 3D medical: aggregating geometric and semantic information simultaneously. in order to extract hierarchical features to deal with objectswith diverse scales. from top to bottom, the resolution is decreasing while the receptive field size is increasing. original resolution of input point cloud through up-sampling layer 
% T-RODNet: nterpolation is used instead of transpose convolution for up-sampling in other radar-based models, which reduces the parameter size and complexity. By the way of skipping connections, the network uses the output of the up-sampled and encoded parts in the channel dimension, and then performs dimensionality reduction through linear layers
% Fig. 2 illustrates the overall structure of T-RODNet, which mainly consists of two parts: encoder and decoder. Overall, T-RODNet adopts a U-shaped frame, which facilitates the extraction of multiscale features.
% 3DCTN: an MLP Head layer is utilized to get the global classification logits, which consists of three linear layers with the batch normalization and ReLU.
% PU: local information is aggregated from both the geometric and feature context of the points, implying their 3D positional relations

\begin{figure*}[t]
    \centering
    \includegraphics[width=0.6\linewidth]{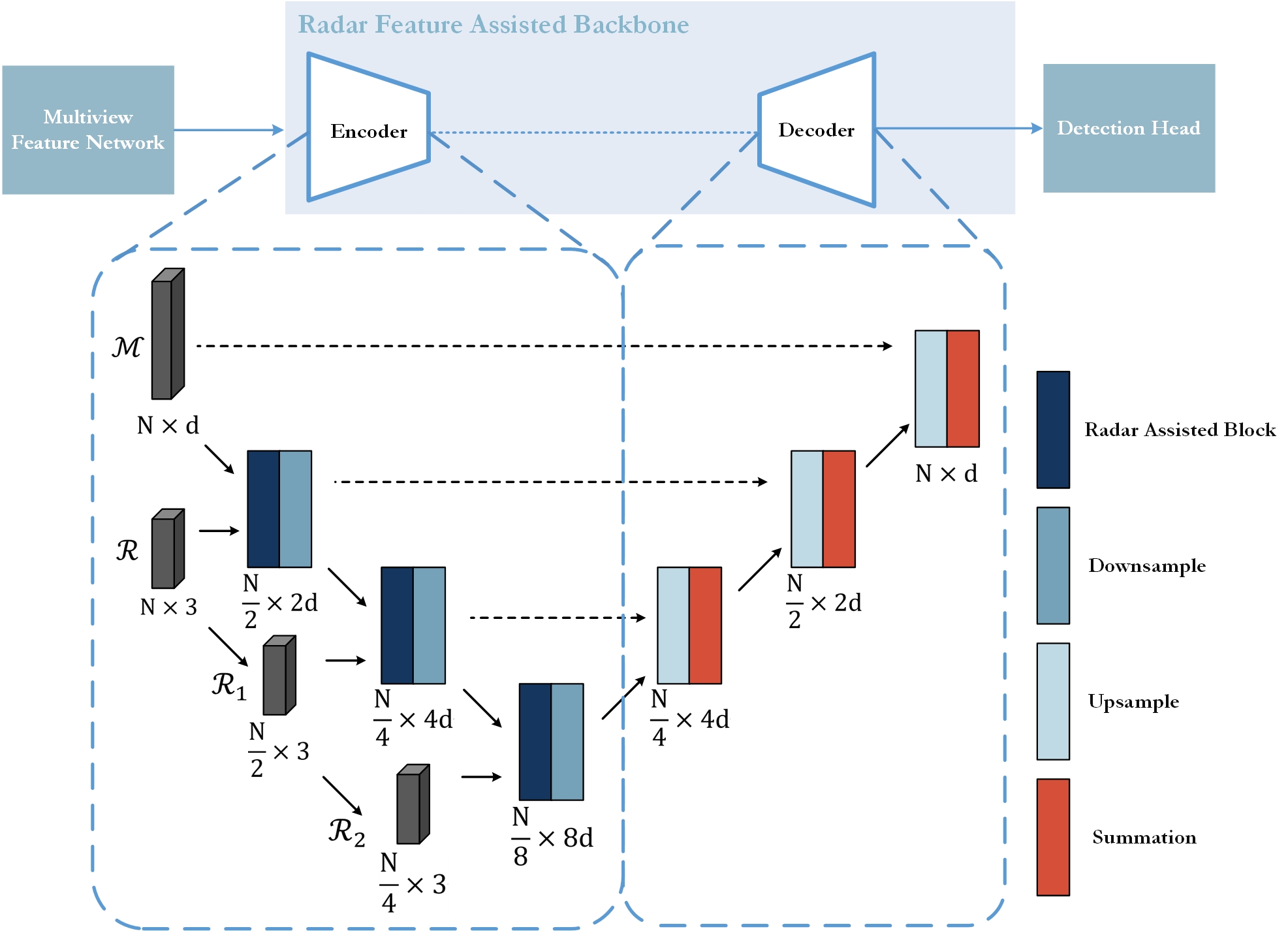}
    \caption{The Radar Feature Assisted backbone.}
    \label{fig_backbone}
\end{figure*}

Radar point clouds exhibit greater sparsity compared to LiDAR, yet they offer additional crucial information in the form of Doppler velocity and reflectivity. However, existing 4D radar-based 3D detectors overlook the significance of these information, failing to fully utilize its potential, further restricting the utilization of radar point clouds. By neglecting to incorporate the essential Doppler and reflectivity data into their frameworks, these methods miss out on valuable insights that could enhance the overall performance of radar-based object detection. Fig.\ref{fig_backbone} illustrates the architecture of our proposed Radar Feature Assisted Backbone, which mainly consists of an encoder and a decoder. The backbone is designed with a U-Net \cite{Unet} structure to extract hierarchical features for identifying objects at multiscale. In the encoder, the resolution decreases while the receptive field increases. In the decoder, the point features are restored back to the original dimension using an upsampling module with a skipped connection. In this paper, we define radar-assisted features as information directly captured by radar, including the absolute Doppler velocity, the relative Doppler velocity to ego-vehicle, and the RCS reflectivity, expressed in Eq.\eqref{R}. The backbone takes the reweighted map $\mathcal{M}_0$ and radar-assisted feature $\mathcal{R}_0$ as input and outputs the processed data into the detection head.

\begin{equation}
\label{R}
\mathcal{R}_0 =  \{{p_n}[v, v_r, RCS]\}^{N-1}_{n=0},\quad \mathcal{R}_0\subseteq \mathbb{R}^{N\times 3}
\end{equation}

\begin{figure*}[htbp]
    \centering
    \subfloat[Radar Assisted Block and Downsample module of the encoder.\label{fig_encoder}]{%
        \includegraphics[width=0.5\linewidth]{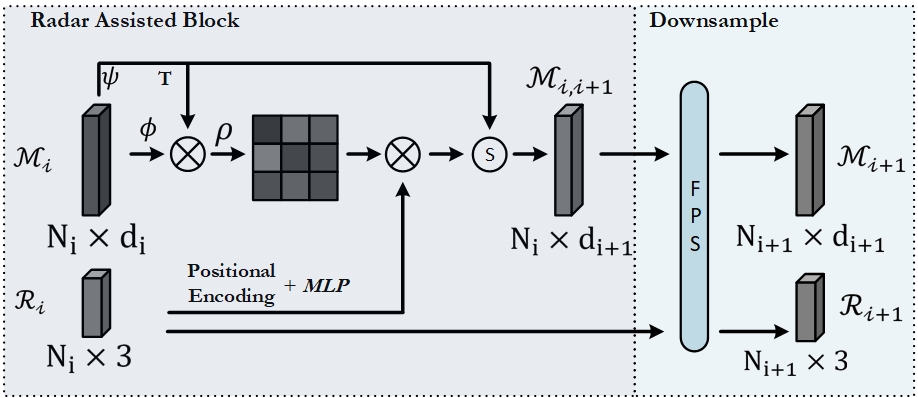}}
    \hfill
    \subfloat[Upsample and Summation module of the decoder.\label{fig_decoder}]{%
        \includegraphics[width=0.33\linewidth]{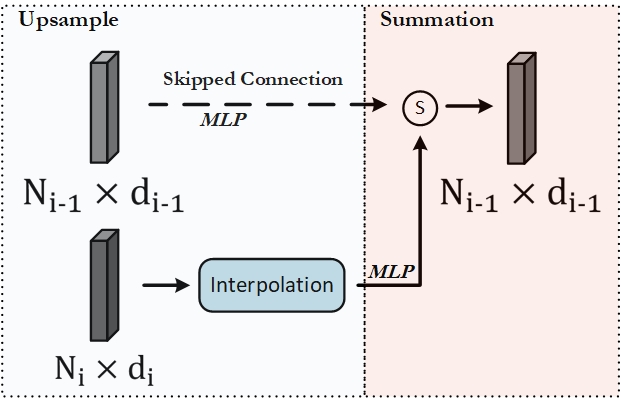}}
    \caption{Details of encoder and decoder structures.}
\end{figure*}

% PT: The input is a set of feature vectors x with associated 3D coordinates p. The point transformer block facilitates information exchange between these localized feature vectors, producing new feature vectors for all data points as its output. The information aggregation adapts both to the content of the feature vectors and their layout in 3D.

% \subsubsection{Encoder Structure} The encoder is composed of three interconnected Radar Assisted Block and Downsample module whose details are shown in Figure~\ref{fig_encoder}. The point-wise feature $\mathcal{M}_i\subseteq \mathbb{R}^{N_i\times d_i}$ and the radar-assisted feature $\mathcal{R}_i\subseteq \mathbb{R}^{N_i\times 3}$ are fed into the Radar Assisted Block where $\mathcal{R}_i$ are firstly processed by same positional encoding in Section.\ref{sec_Positional_Map_Generation}. Then matrix multiplication is performed with a residual connection to output learned feature with $d_{i+1}=2d_i$ dimension. Finally, the farthest point sampling strategy is adopted to downsample all features for enlarging receptive fields. Notably, the radar-assisted feature $\mathcal{R}_i$ is associated with the 3D coordinates of their point clouds and is fused with the point-wise $\mathcal{M}_i$ to enhance the information exchange at the feature vector level in a trainable manner. The Radar Assisted Block aggregates both the content of the feature vectors of $\mathcal{R}_i$ and $\mathcal{M}_i$ with their 3D position. 
\subsubsection{Encoder Structure} The encoder is composed of three interconnected Radar Assisted Blocks and Downsample modules whose details are shown in Fig.\ref{fig_encoder}. Each module takes in the point-wise feature $\mathcal{M}_i\subseteq \mathbb{R}^{N_i\times d_i}$ and the radar-assisted feature $\mathcal{R}_i\subseteq \mathbb{R}^{N_i\times 3}$ as inputs.  $\mathcal{R}i$ is firstly processed using the same positional encoding method described in Section.\ref{sec_Positional_Map_Generation}. Then, matrix multiplication is performed with a residual connection to output learned feature with $d_{i+1}=2d_i$ dimensions. Finally, the farthest point sampling strategy is adopted to downsample all features and enlarge their receptive fields. Notably, $\mathcal{R}_i$ is associated with the corresponding 3D coordinates and is fused with the point-wise $\mathcal{M}_i$ to enhance the information exchange at the feature vector level in a trainable manner. The Radar Assisted Block aggregates both the content of the feature vectors of $\mathcal{R}_i$ and $\mathcal{M}_i$ with their 3D positions.

\begin{equation}\begin{gathered}
\label{encoder}
\mathcal{M}_{i,i+1} = \rho(\phi(\mathcal{M}_i)\psi(\mathcal{M}_i)^T)\times \theta(\mathcal{R}_i, P_i^k, ||P_i-P^k||) + \phi(\mathcal{M}_i)\\
\mathcal{M}_{i+1} = \text{FPS}(\mathcal{M}_{i,i+1}) \\
\mathcal{R}_{i+1} = \text{FPS}(\mathcal{R}_i)
\end{gathered}
\end{equation}
where $\phi,\psi,\theta$ share the same definition as $MLP$ and $\rho$ is the sigmoid function.

\subsubsection{Decoder Structure}

The decoder is a symmetrical structure of the encoder, as shown in Fig.\ref{fig_decoder}. Consecutive stages in the decoder are connected by Upsample and Summation modules. The objective of the decoder is to restore features from the downsampled $\mathcal{M}_i$ to their previous shape as $\mathcal{M}_{i-1}\subseteq \mathbb{R}^{N_{i-1}\times d_{i-1}}$. This is achieved by the up-sampling layer through interpolation in point size dimension and a $MLP$ in feature dimension. Then, the interpolated features from the preceding decoder stage are summed with the features from the corresponding encoder stage via a skipped connection.

\textbf{Discussion. }Our proposed Radar Feature Assisted Backbone distinguishes itself from existing approaches by effectively harnessing crucial information, specifically the Doppler velocity and reflectivity measured by radar. This innovative approach significantly enhances the performance of feature extraction and strengthens the high-level feature $\mathcal{M}_i$ by incorporating it with the low-level feature $\mathcal{R}_i$ through additional position encoding. One of the remarkable features of radar is its ability to measure reflectivity and Doppler velocity, as well as infer related Doppler velocity as the low-level feature $\mathcal{R}_i$. Surprisingly, this critical information has not been leveraged in 4D radar-based 3D object detection before. To showcase the effectiveness of our approach, we conduct an ablation study that clearly demonstrates the benefits of incorporating this previously unexploited information.

\subsection{Detection Head and Loss Function}\label{sec_head&loss}
We implement the identical SSD\cite{SSD} detection head, which is commonly embedded in SECOND\cite{second}, PointPillars, PillarOD and MVF. The localization loss $\mathcal{L}_{loc}$ and the angle loss $\mathcal{L}_{dir}$ are computed using a Smooth L1 loss function that is regressed between the ground truth boxes $(x^{gt}, y^{gt}, z^{gt}, w^{gt}, l^{gt}, h^{gt}, \theta^{gt})$ and anchors $(x^{pre}, y^{pre}, z^{pre}, w^{pre}, l^{pre}, h^{pre}, \theta^{pre})$ as Eq.\eqref{loss_loc} where the diagonal of the base of the anchor box is referred as $d^{pre}=\sqrt{(w^{pre})^2+(l^{pre})^2}$. $\mathcal{L}_{dir}$ is proven by \cite{second} to effectively discretize directions of flipped boxes. The classification loss $\mathcal{L}_{cls}$ is calculated by the focal loss, written as Eq.\eqref{loss_cls} where the predicted classification probability is revealed as $p^{pre}$.

\begin{equation}\label{loss_loc}
\begin{gathered}
\mathcal{L}_{loc} = \sum_{b \in (x,y,z,w,l,h)} \text{Smooth L1}[\Delta b]\\
\Delta x=\frac{x^{pre}-x^{gt}}{d^{pre}},\Delta y=\frac{y^{pre}-y^{gt}}{d^{pre}},\Delta z=\frac{z^{pre}-z^{gt}}{d^{pre}}\\
\Delta w=\log\frac{w^{gt}}{w^{pre}},\Delta l=\log\frac{l^{gt}}{l^{pre}},\Delta h=\log\frac{h^{gt}}{h^{pre}}\\
\mathcal{L}_{dir}=\text{Smooth L1}[sin(\theta^{pre}-\theta^{gt})]
\end{gathered}
\end{equation}  

\begin{equation}\label{loss_cls}
\begin{gathered}
\mathcal{L}_{cls} = - \alpha_{pre}(1-p^{pre})^\gamma \ln p^{pre},\quad \alpha_{pre}=0.25, \gamma=2
\end{gathered}
\end{equation}  

By combing all losses discussed in Eq.\eqref{ax_loss},Eq.\eqref{loss_loc} and Eq.\eqref{loss_cls}, total loss function is asddressed as Eq.\eqref{loss_total} where $ \alpha_{1}= \alpha_{2}=1,  \alpha_{3}=2, \alpha_{4}=0.2$.

\begin{equation}\label{loss_total}
\begin{gathered}
\mathcal{L}_{total} = \alpha_{1}\mathcal{L}_{cls} + \alpha_{2}\mathcal{L}_{ax} +\alpha_{3}\mathcal{L}_{loc}+\alpha_{4}\mathcal{L}_{dir}
\end{gathered}
\end{equation}

\section{Experiments}\label{sec_experiement}
\subsection{Dataset}\label{sec_dataset}

We evaluate our proposed method on Astyx and VoD datasets to demonstrate its effectiveness and efficiency. Astyx is composed of 546 frames with over 3000 3D object annotations in seven distinct classes: cars, buses, humans, bicycles, motorcycles, trucks, and trailers, with cars being the majority. Due to the imbalanced distribution, we focus solely on the car class for training and evaluation and adopt a 75\% training and 25\% validation split ratio following \cite{RPFA-Net}. VoD ensures a proportional and reasonable distribution of multi-class annotations, including cars, pedestrians, and cyclists, and is officially divided into training, validation, and testing sets at proportions of 59\%, 15\%, and 26\%. 

% Additionally, VoD experiments with accumulated radar point cloud data by transforming and stacking the preceding 3 and 5 frames with appropriate variant time IDs and combining them into one frame before inputting them into the network, consequently improving detection precision.

\subsection{Evaluation}

Astyx uses average precision (AP) as an accuracy metric for the car class with an intersection over union (IoU) threshold of 0.5. Furthermore, it classifies objects into three difficulty levels - easy, moderate, and hard - based on their visibility and the degree of occlusion. VoD dataset provides average precision (AP) and average orientation similarity (AOS) as two evaluation metrics. Specifically, the 3D IoU threshold for the car class is set at 0.5, while the thresholds for pedestrians and cyclists are both established at 0.25. Additionally, VoD organizes the detection process into two levels, (a) all areas within field-of-view, (b) the driving corridor at $[0<x<25m, -4m<y<4m]$.

\subsection{Implementation Details}\label{sec_implementation}
{\bf{Network Parameter.}} Several key parameters need to be clarified to understand the spatial resolution of BEV pillar projection for $x$, $y$, $z$ axes and cylindrical resolution for $\rho$, $\phi$, $z^{\prime}$ axes with the detection range in Tab.\ref{tab_param}

% of the two datasets, Astyx and VoD. For Astyx dataset, the spatial resolution of BEV pillar projection for $x$, $y$, $z$ axes is $0.16m$, $0.16m$, and $4m$, whereas for VoD dataset, it is $0.16m$, $0.16m$, and $5m$. The detection ranges of point clouds are as follows: for Astyx, the $x$, $y$, $z$ axes range from $[0, 99.84m]$, $[-39.68m, 39.68m]$, and $[-3m, 1m]$, while for VoD, they range from $[0, 51.2m]$, $[-25.6m, 25.6m]$, and $[-3m, 2m]$. In cylindrical coordinates, the Astyx dataset has resolutions of $100.6m$, $\pi/1280$, and $0.04m$ for $\rho$, $\phi$, $z^{\prime}$ axes, with a range of $[0, 100.6m], [0, \pi], [-3m, 1m]$. In contrast, the VoD dataset has resolutions of $72.4m$, $\pi/1280$, and $0.05m$ for $\rho$, $\phi$, $z^{\prime}$ axes, and a range of $[0, 72.4m], [0, \pi], [-3m, 2m]$.

\begin{table}[t]
\captionsetup{justification=centering}
\caption{Network parameters setting for two datasets.}
\centering
\resizebox{\columnwidth}{!}{%
\label{tab_param}
\begin{tabular}{c|cccccc|ccc|ccc}
\toprule
 Dataset & \multicolumn{6}{c|}{Detection range} & \multicolumn{3}{c|}{BEV Resolution} & \multicolumn{3}{c}{Cylindrical Resolution} \\ 
 & $x$ & $y$ & $z$ & $\rho$ & $\phi$ & $z^{\prime}$ & $x$ & $y$ & $z$ & $\rho$ & $\phi$ & $z^{\prime}$  \\ \hline
Astyx & $[0, 99.84m]$ & $[-39.68m, 39.68m]$ & $[-3m, 1m]$ & $[0, 100.6m]$ & $[0, \pi]$ & $[-3m, 1m]$ & $0.16m$ & $0.16m$ & $4m$ & $100.6m$ & $\pi/1280$ & $0.04m$ \\
VoD & $[0, 51.2m]$ & $[-25.6m, 25.6m]$ & $[-3m, 2m]$ & $[0, 72.4m]$ & $[0, \pi]$ & $[-3m, 2m]$ & $0.16m$ & $0.16m$ & $5m$ & $72.4m$ & $\pi/1280$ & $0.05m$ \\ \bottomrule
\end{tabular}%
}
\end{table}

% For cylinder pillar project described in Section \ref{sec_cylinder_project}, the resolution in $\rho$, $\phi$, $z^{\prime}$ axis are $72.4m$, $\pi/1280$ and $0.05m$ with the range $[0, 72.4m], [0, \pi], [-3m, 2m]$ in cylinder coordinate.

% The radar feature-assisted backbone is adapted from PointPillars by substituting our ResNet blocks for the convolution blocks in the Eq.\eqref{backbone1} and addressed in Section \ref{sec_backbone}.

% {\bf{Settings.}} The settings of matching strategy and generated anchors for \textit{car}, \textit{pedestrian} and \textit{cyclist} are identical with PointPillars and VoxelNet \cite{voxelnet}.The only parameter that has been changed is the maximum number of points that can be found on a single pillar, which has been set to $10$ due to the sparse nature of the radar point clouds. 

{\bf{Data Augmentation.}} Due to the fact that the Doppler velocity is relative to the vehicle's motion, only a limited data augmentation technique can be applied to radar. Therefore, we only seek to randomly flip the point cloud around the $x$ axis and scale the point clouds by a factor within $[0.95, 1.05]$. For radar point clouds, translation and rotation of components are not adopted.

{\bf{Training.}} To train 80 epochs, we adopt the ADAM optimization with $0.003$ initial learning rate and $0.01$ weight decay rate on $4$ Nvidia RTX 2080Ti GPUs with a batch size of 8.

% \begin{table*}[ht]
% % 
% \centering
% \captionsetup{justification=centering}
% \caption{Comparison of different grid size settings for BEV and cylinder view: mean average precision (mAP) and mean average orientation similarity (mAOS) in \textit{all area} and \textit{driving corridor}.}
% \label{tab_voxel_2}
% \begin{tabular}{cccccc}
% \toprule
% Grid Size & \multicolumn{2}{c}{All Area (\%)} & \multicolumn{2}{c}{Driving Corridor (\%)} & \multirow{2}{*}{FPS (\textit{Hz})} \\
% $BEV-cyl$ & mAP & mAOS & mAP & mAOS & \\ \hline
% $BEV8-cyl4$ & 38.42 & 29.33 & 63.38 & 53.46 &  26.04 \\
% $BEV8-cyl5$ & 37.89 & 30.02 & 62.93 & 55.69 & 28.39 \\
% $BEV8-cyl6$ & 39.51 & 30.63 & 63.67 & 53.38 & 29.95 \\
% $BEV16-cyl4$ & \textbf{40.18} & \textbf{31.75} & 64.36 & 56.41 & 39.12 \\ \hline
% $BEV16-cyl5$ & 39.42 & 31.58 & \textbf{64.38} & \textbf{57.01} & 45.11 \\ \hline
% $BEV16-cyl6$ & 39.45 & 31.40 & 64.23 & 56.42 & 49.38 \\
% $BEV32-cyl4$ & 39.34 & 29.45 & 64.29 & 55.78 & 44.13 \\
% $BEV32-cyl5$ & 39.06 & 31.42 & 64.09 & 56.06 & 51.20 \\
% $BEV32-cyl6$ & 38.51 & 30.45 & 63.64 & 53.50 & 56.32 \\ \bottomrule
% \end{tabular}
% \end{table*}

\begin{table*}[t]
\centering
\captionsetup{justification=centering}
\caption{Comparative experiments on Astyx. The mean average precisions for \textit{car} class are based on 3D and BEV IoU respectively.}
\label{result_astyx}
\begin{tabular}{ccccccc}
\toprule
\multirow{2}{*}{Methods} & \multicolumn{3}{c}{3D (\%)} & \multicolumn{3}{c}{BEV (\%)} \\
 & Easy & Moderate & Hard & Easy & Moderate & Hard \\ \hline
PointRCNN \cite{PointRCNN} & 12.23 & 9.10 & 9.10 & 14.95 & 13.82 & 13.89 \\
SECOND \cite{second} & 24.11 & 18.50 & 17.77 & 41.25 & 30.58 & 29.33 \\
H\textsuperscript{2}3DR-CNN \cite{hollow_3D} & 26.21 & 20.12 & 21.62 & 45.23 & 32.58 & 35.54 \\
PV-RCNN \cite{PV-RCNN} & 28.21 & 22.29 & 20.40 & 46.62 & 35.10 & 33.67 \\
PointPillars \cite{pointpillar} & 30.14 & 24.06 & 21.91 & 45.66 & 36.71 & 35.30 \\
RPFA-Net \cite{RPFA-Net} & 38.85 & 32.19 & 30.57 & 50.42 & 42.23 & 40.96 \\ \hline
MVFAN (\textit{ours}) & \textbf{45.60} & \textbf{39.52} & \textbf{38.53} & \textbf{58.68} & \textbf{50.24} & \textbf{46.56} \\ \bottomrule
\end{tabular}%

\end{table*}

\begin{table*}[t]
\centering
\captionsetup{justification=centering}
\caption{Comparative experiments on VoD: Average precision for \textit{car}, \textit{pedestrian}, \textit{cyclist}, mean average precision (mAP) and mean average orientation similarity (mAOS).}
\label{tab_method}
\resizebox{\columnwidth}{!}{%
\begin{tabular}{cccccccccccc}
\toprule
\multirow{2}{*}{Methods} & \multicolumn{5}{c}{All   Area (\%)} & \multicolumn{5}{c}{Driving   Corridor (\%)} & \multirow{2}{*}{FPS (\textit{Hz})} \\
 & Car & Ped. & Cyc. & mAP & mAOS & Car & Ped. & Cyc. & mAP & mAOS & \\ \hline
PointRCNN \cite{PointRCNN} & 15.98 & 24.52 & 43.14 & 27.88 & 23.40 & 33.60 & 33.73 & 60.32 & 42.55 & 37.73 & 14.31 \\
PV-RCNN \cite{PV-RCNN} & 31.94 & 27.19 & 57.01 & 38.05 &24.12 & 67.76 & 38.03 & 76.99 & 62.66 & 40.79 & 23.49\\
H\textsuperscript{2}3DR-CNN \cite{hollow_3D} & 25.90 & 14.88 & 34.07 & 24.95 & 16.91 & 61.89 & 23.48 & 53.95 & 46.44 & 35.42 & 43.59 \\
SECOND \cite{second} & 32.35 & 24.49 & 51.44 & 36.10 & 27.58 & 67.98 & 35.45 & 72.30 & 59.18 & 50.92 & 80.87 \\
PointPillars \cite{pointpillar} & 32.54 & 26.54 & 55.11 & 38.09 & 30.10 & 68.78 & 36.02 & 80.63 & 62.58 & 53.96 & 157.50 \\ 
RPFA-Net \cite{RPFA-Net} & 33.45 & 26.42 & 56.34 & 38.75 & 31.12 & 68.68 & 34.25 & 80.36 & 62.44 & 52.36 & 100.23 \\ \hline
MVFAN(\textit{ours}) & \textbf{34.05} & \textbf{27.27} & \textbf{57.14} & \textbf{39.42} & \textbf{31.58} & \textbf{69.81} & \textbf{38.65} & \textbf{84.87} & \textbf{64.38} & \textbf{57.01} & 45.11 \\ \bottomrule
\end{tabular}
}
\end{table*}

\subsection{Experiment Results}\label{sec_result}
% Because VoD dataset was released online several months ago, few 4D radar detection frameworks have been developed. Therefore, 

% We compare \textit{MVFAN} with state-of-the-art detectors for comparison, such as PointRCNN \cite{PointRCNN}, PV-RCNN \cite{PV-RCNN}, H\textsuperscript{2}3DR-CNN \cite{hollow_3D}, SECOND \cite{second} and PointPillars \cite{pointpillar} with superior performance in KITTI\cite{kitti} dataset. We re-train these frameworks on Astyx and VoD from scratch. 
We compare \textit{MVFAN} with state-of-the-art point cloud detectors for comparison, PointRCNN \cite{PointRCNN}, PV-RCNN \cite{PV-RCNN}, H\textsuperscript{2}3DR-CNN \cite{hollow_3D}, SECOND \cite{second}, PointPillars \cite{pointpillar} and RPFA-Net \cite{RPFA-Net}. We re-train these frameworks on Astyx and VoD from scratch. Since our Multi-View Feature Extraction module is inspired by MVF \cite{MVF} which exploits variant views, we are supposed to re-implement it. Nevertheless, MVF has not been released, so its final result is unapproachable. As elaborated in Tab.\ref{result_astyx} and \ref{tab_method}, our \textit{MVFAN} beats all of these 3D detectors. 

% Even the most competitive architecture PV-RCNN cannot match the accuracy of \textit{MVFAN}.

% which reaffirms our concern that directly employing LiDAR-based detectors on the radar is inappropriate. 

\subsection{Ablation Study}\label{sec_ablatio}
We demonstrate the effectiveness of two major components of our model, Multi-view Feature Extraction (\textit{MV}) and Radar Feature Assisted Backbone (\textit{FA}). Since our framework is an adaptation of state-of-the-art PointPillars (\textit{PP}), we apply it as the baseline in our experiments. Tab.\ref{tab_ab} underlines the validation of a few combinations of \textit{MV} and \textit{FA} on the detection outcome. 

\textbf{(a)} Firstly, we re-train PP. As radar point clouds consist of spatial coordinates, Doppler velocity and RCS reflectivity, the input channel of PointNet $(D=9, P, N)$ is expanded to $(D=11, P, N)$ as stated in VoD.

% \textit{FA} is proved to have a slight effect on detecting \textit{cars} when compared to the baseline, (from $32.54$ to $32.55$ in all area and from $68.78$ to $68.84$ in driving corridor) whereas it makes a significant difference on detecting small moving object such as \textit{pedestrians} (increasing from $26.54$ to $27.23$ in all area and from  $36.02$ to $36.69$ in driving corridor) and \textit{cyclists} (increasing from $55.11$ to $55.85$ in all area and from $80.63$ to $81.57$ in driving corridor). Since bounding boxes of \textit{pedestrians} and \textit{cyclists} only include a limited number of radar points, the Doppler motions and observed RCS reflectivity allow our network to identify these objects the best.

\textbf{(b)} Secondly, we only utilize \textit{MV} on \textit{PP}. Instead of feeding solely BEV features into the backbone as PointPillars, retrieved dense point-wise cylinder features and  point-wise BEV features are concatenated followed by weight map multiplication. It validates that \textit{MV} largely improves the performance of all three characters. Focusing on the driving corridor, which VoD identifies as a safety-critical region, \textit{MV} also highly magnifies the dense representation for the detector in order to distinguish objects from the background, particularly small volume items, such as \textit{pedestrians} and \textit{cyclists}, whose contours reveal a drastic deviation in cylinder perspective. When objects in all areas are examined, \textit{MV} refines the performance and escalates the indicator by $0.62\%$, $0.70\%$ and $0.39\%$. The advancement of accuracy highlights the essence and capacity of the multi-view sampling scheme and reweighting of the foreground and background points for better maintaining 3D contextual information and morphological characteristics more effectively. 

\begin{table*}[t]
\centering
\captionsetup{justification=centering}
\caption{Ablation study: Effectiveness of \textit{MV} and \textit{FA}.}
\label{tab_ab}
\resizebox{\columnwidth}{!}{%
\begin{tabular}{lccccccccccc}
\toprule
\multirow{2}{*}{Methods} & \multicolumn{5}{c}{All   Area (\%)} & \multicolumn{5}{c}{Driving   Corridor (\%)} & \multirow{2}{*}{FPS (\textit{Hz})} \\
 & Car & Ped. & Cyc. & mAP & mAOS & Car & Ped. & Cyclist & mAP & mAOS & \\ \hline
(a)\ \textit{PP} & 32.54 & 26.54 & 55.11 & 38.09 & 30.10 & 68.78 & 36.02 & 80.63 & 62.58 & 53.96 & 157.50 \\
(b)\ \textit{PP}+\textit{MV} & 33.16 & 27.24 & 55.50 &38.67 & 30.65 & 69.71 & 38.62 & 83.91 & 64.22 & 56.32 & 145.56\\
(b)\ \textit{PP}+\textit{FA} & 32.55 & 27.23 & 55.85 & 38.51 & 30.73 & 68.84 & 36.69 & 81.57 & 62.37 & 53.79 & 45.77 \\
(d)\ \textit{MVFAN} & \textbf{34.05} & \textbf{27.27} & \textbf{57.14} & \textbf{39.42} & \textbf{31.58} & \textbf{69.81} & \textbf{38.65} & \textbf{84.87} & \textbf{64.38} & \textbf{57.01} & 45.11 \\ \bottomrule
% (e)\ \textit{MVFAN 3frames} & 40.01 & \textbf{36.09} & 66.28 & 46.80 & 38.34 & 71.46 & 42.07 & \textbf{87.92} & 67.15 & 61.41 & 45.01 \\ 
% (f)\ \textit{MVFAN 5frames} & \textbf{41.27} & 35.84 & \textbf{66.98} & 47.03 & 38.07 & \textbf{71.49} & \textbf{43.49} & 87.48 & 67.45 & 61.76 & 44.95 \\ \bottomrule
\end{tabular}
}
\end{table*}

\begin{table*}[t]
\centering
\captionsetup{justification=centering}
\caption{Ablation study: Effectiveness of Positional Map.}
\label{tab_weight}
\resizebox{\columnwidth}{!}{%
\begin{tabular}{lccccccccccc}
\toprule
\multirow{2}{*}{Methods} & \multicolumn{5}{c}{All   Area (\%)} & \multicolumn{5}{c}{Driving   Corridor (\%)} & \multirow{2}{*}{FPS (\textit{Hz})} \\
 & Car & Ped. & Cyc. & mAP & mAOS & Car & Ped. & Cyclist & mAP & mAOS & \\ \hline
(a)\ \textit{PP}+\textit{MV w/o} & 32.66 & 26.82 & 55.23 & 38.25 & 30.32 & 68.94 & 37.42 & 82.54 & 63.20 & 54.32 & 150.66\\
(b)\ \textit{PP}+\textit{MV w} & 33.16 & 27.24 & 55.50 &38.67 & 30.65 & 69.71 & 38.62 & 83.91 & 64.22 & 56.32 & 145.56\\
(c)\ \textit{MVFAN w/o} & 33.98 & 27.25 & 56.86 & 39.22 & 31.23 & 69.75 & 38.63 & 84.52 & 64.31 & 56.88 & 49.42 \\
(d)\ \textit{MVFAN w} & \textbf{34.05} & \textbf{27.27} & \textbf{57.14} & \textbf{39.42} & \textbf{31.58} & \textbf{69.81} & \textbf{38.65} & \textbf{84.87} & \textbf{64.38} & \textbf{57.01} & 45.11 \\ \bottomrule
\end{tabular}
}
\end{table*}

\textbf{(c)} Next, we only replace the backbone of \textit{PP} with \textit{FA}. The average precision of this model is upgraded by incorporating additional radar information by matrix operation at a minimal cost in terms of inference time, which preserves the Doppler and RCS reflectivity of radar point clouds. \textit{FA} is proved to have a slight effect on detecting \textit{cars}, whereas it makes a significant difference on detecting small moving objects such as \textit{pedestrians} and \textit{cyclists}. In scenarios where the bounding boxes of \textit{pedestrians} and \textit{cyclists} encompass only a limited number of radar points, relying solely on the 3D geometric information from sparse point clouds may not yield optimal results. Instead, the Doppler motions and observed RCS reflectivity play a crucial role in accurately identifying these objects. By incorporating the Doppler motions and RCS reflectivity into our network, we are able to achieve superior object identification performance. This outcome effectively demonstrates the effectiveness and importance of leveraging the Doppler and RCS reflectivity information.

% For all three classes, \textit{cars}, \textit{pedestrians}, \textit{cyclists} and mAP, \textit{MV} uplifts the average precision by $1.4\%$, $7.2\%$, $4.1\%$ and $2.6\%$ in driving corridor, and enhances the angular precision by $4.1\%$ in terms of mAOS. When objects in all areas within $50m$ and camera's FOV are examined, multi-view sampling paradigm refines the perform and escalates the indicator by $1.9\%$, $3.0\%$, $0.7\%$, $1.5\%$ and $1.5\%$. The advancement of accuracy highlights the essence and capacity of multi-view sampling scheme for better maintaining 3D contextual information and morphological characteristics more effectively. Notably, while our \textit{MV} introduces ResNet to extract semantic feature map from two input streams, it ameliorates the performance at the expense of increasing computation, but yet is still able to execute in real-time.

\textbf{(d)} Moreover, our proposed framework \textit{MVFAN}, a combination of \textit{MV} and \textit{FA}, exhibits a high-ranking detection precision. The average precision for all three classes is further upgraded compared to when either \textit{MV} or \textit{FA} is independently applied on the baseline. Overall, \textit{MVFAN} boosts the mAP by $3.4\%$ in all area and $2.8\%$ in the driving corridor and elevates the angular accuracy, mAOS, by $4.9\%$ in all area and $5.6\%$ in the driving corridor. Impressively, these two procedures function concurrently and independently with \textit{MV} augmenting the point-wise semantic features by leveraging two input streams to reward the sparsity of radar point clouds whereas \textit{FA} amplifying the fused feature map after \textit{MV} module by adopting additional radar information. Remarkably, the promotion raised by \textit{MVFAN} lies in the emphasis on multi-view feature extraction with reweighing points and rational enforcement of radar Doppler and reflectivity.

Tab.\ref{tab_weight} confirms the effectiveness of our proposed Positional Map in improving the performance of both PointPillars and our \textit{MVFAN}. The inclusion of the Positional Map leads to higher mAP and mAOS scores compared to the baseline methods (\textit{PP}+\textit{MV w/o} and \textit{MVFAN w/o}). It enhances the detection of cars, pedestrians, and cyclists, with improved percentages in those categories in both all area and the driving corridor scenario. These results validate the effectiveness of the Positional Map in enhancing the overall object detection performance.

% \textbf{(e\&f)} Furthermore, we train our model examining accumulated radar points of prior $3$ and $5$ frames as \textit{MVFAN 3frames} and \textit{MVFAN 5frames} to demonstrate that mAP and mAOS increase considerably by adding additional input from prior frames rather than inputting a single frame.
\begin{table*}[htbp]
\centering
\captionsetup{justification=centering}
\caption{Comparison of different grid size settings for BEV and cylinder view.}
\label{tab_voxe_1}
\resizebox{\columnwidth}{!}{%
\begin{tabular}{cccccccccccc}
\toprule
Grid Size & \multicolumn{5}{c}{All Area (\%)} & \multicolumn{5}{c}{Driving Corridor (\%)}& \multirow{2}{*}{FPS (\textit{Hz})}\\
$BEV-cyl$ & Car & Ped. & Cyclist & mAP & mAOS & Car & Ped. & Cyclist & mAP & mAOS & \\ \hline
$BEV8-cyl4$ & 32.74 & 26.54 & 55.99 & 38.42 & 29.33 & 69.35 & 36.53 & 84.29 & 63.38 & 53.46 & 26.04 \\
$BEV8-cyl5$ & 32.18 & 26.56 & 54.96 & 37.89 & 30.03 & 69.51 & 38.24 & 83.06 & 62.93 & 55.69 & 28.39\\
$BEV8-cyl6$ & 32.27 & 26.79 & 54.82 & 39.51 & 30.63 & 68.55 & 38.32 & 84.11 & 63.67 & 53.38 & 29.95 \\
$BEV16-cyl4$ & 34.04 & \textbf{27.56} & \textbf{57.54} &\textbf{40.18} & \textbf{31.75} & 69.41 & \textbf{38.69} & \textbf{85.34} & 64.36 & 56.41 & 39.12 \\ \hline
$BEV16-cyl5$ & \textbf{34.05} & 27.27 & 57.14 & 39.42 & 31.58 & \textbf{69.81} & 38.65 & 84.87 & \textbf{64.48} & \textbf{57.01} & 45.11\\ \hline
$BEV16-cyl6$ & 33.76 & 27.42 & 57.16 & 39.45 & 31.40 & 69.80 & 37.67 & 84.21 & 64.23 & 56.42 & 49.38 \\
$BEV32-cyl4$ & 33.27 & 25.89 & 56.86 & 39.34 & 29.45 & 69.80 & 38.29 & 84.69 & 64.29 & 55.78 & 44.13 \\
$BEV32-cyl5$ & 33.18 & 26.44 & 57.53 & 39.06 & 31.42 & 69.51 & 38.03 & 84.72 & 64.09 & 56.06 & 51.20 \\
$BEV32-cyl6$ & 32.06 & 26.32 & 57.12 & 38.51 & 30.45 & 69.38 & 37.11 & 84.42 & 63.64 & 53.50 & 56.32 \\ \bottomrule
\end{tabular}
}
\end{table*}

To investigate the effect of parameter settings, we deploy various combinations of partition parameters. For BEV pillar projection, we implement three types of grid size $BEV8:[0.08m, 0.08m, 5m]$, $BEV16:[0.16m, 0.16m, 5m]$ and $BEV32:[0.32m, 0.32m, 5m]$ and for cylinder pillar projection, we test three types of grid size $cyl4:[72.4m, \pi/1600, 0.04m]$, $cyl5:[72.4m, \pi/1280, 0.05m]$ and $cyl6:[72.4m, \pi/1000, 0.0625m]$. There are $9$ experiments conducted in total as presented in Tab.\ref{tab_voxe_1}. By partitioning the BEV with course grids, $BEV32$, each pillar has a greater number of point clouds but also incorporates more noise reflected by environments as opposed to objects. Although a finer grid $BEV8$ brings a higher spatial resolution to separate point clouds, the sparsity of radar introduces more vacant cells and necessitates additional computation. For the partition of cylinder projection, we have already established an exceptional resolution for dividing grid cells, for $cyl4, cyl5, cyl6$ with the same $BEV$ grid size, the performances are only a little altered, on the contrary, the inference speed is largely impacted.

\begin{figure*}[t]
    \centering
    \includegraphics[width=0.9\linewidth]{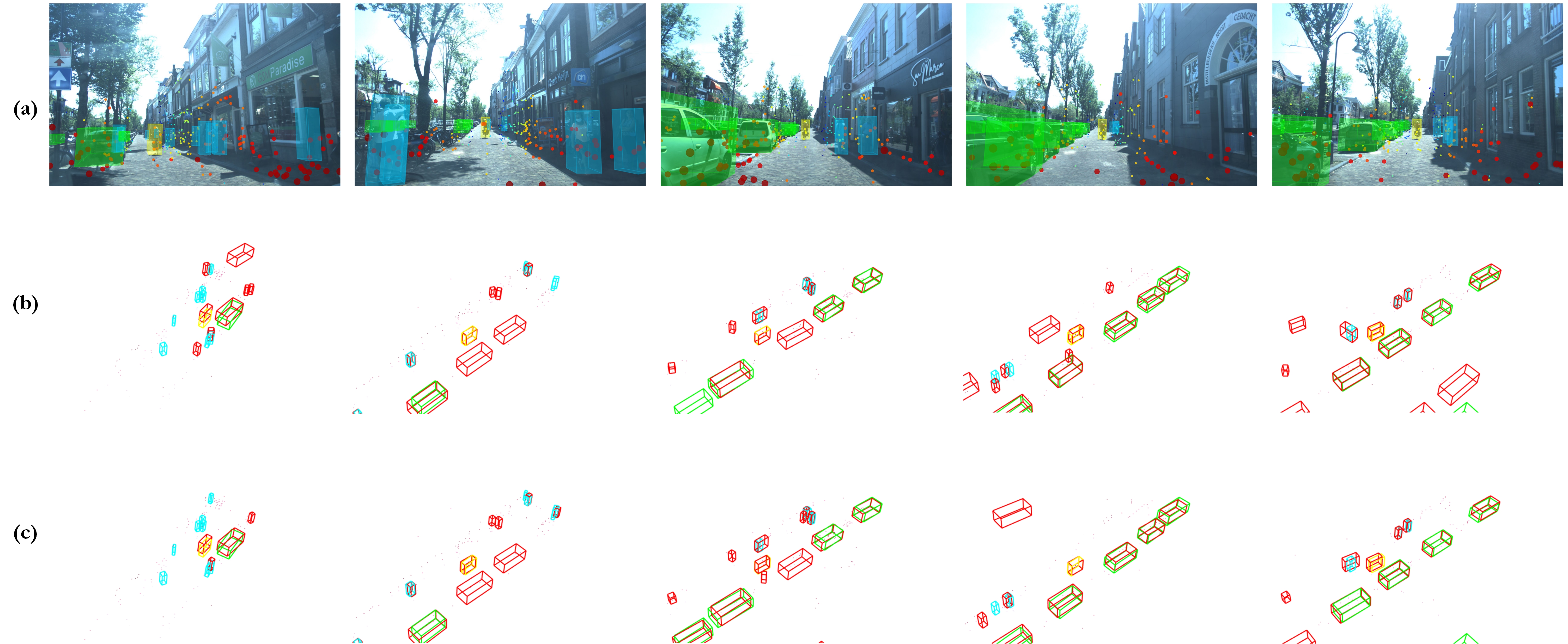}
    \caption{Qualitative results on the VoD dataset.}
    \label{fig_qualitative}
\end{figure*}

\begin{figure*}[t]
    \centering
    \includegraphics[width=0.9\linewidth]{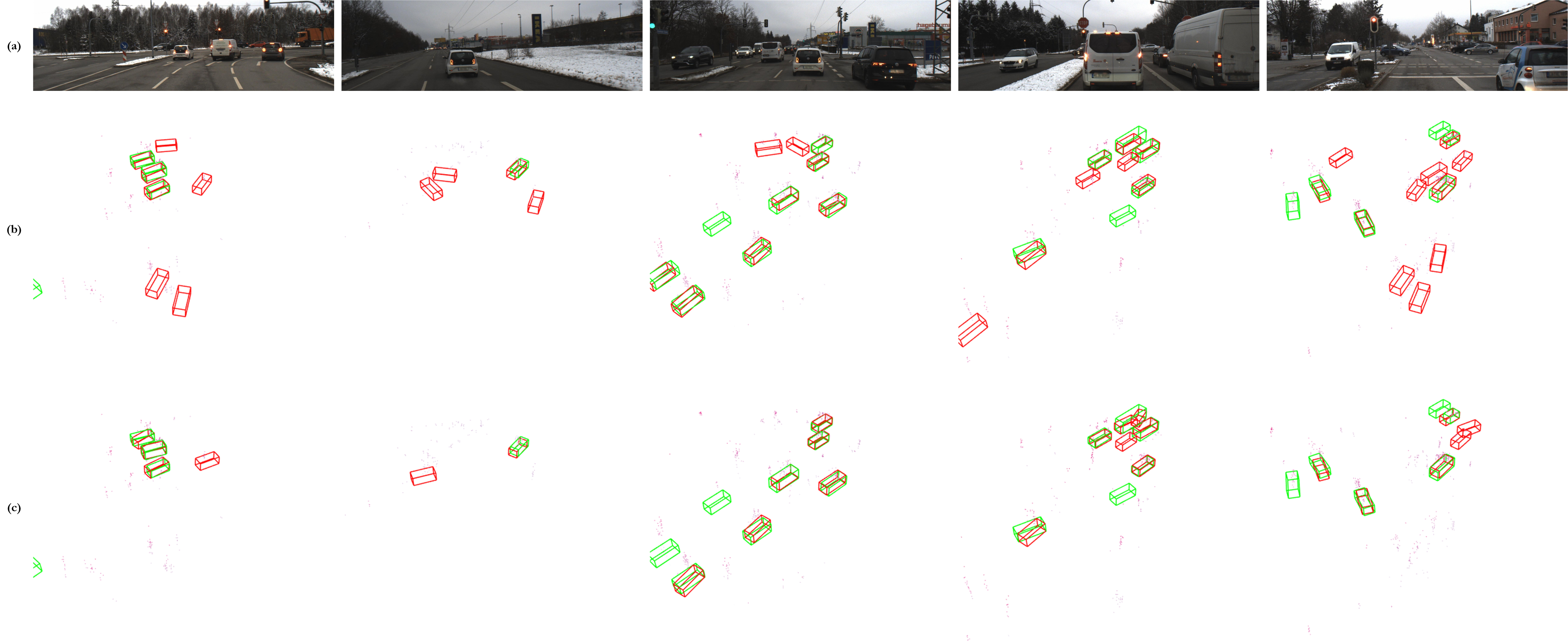}
    \caption{Qualitative results on the Astyx dataset.}
    \label{fig_qualitative_astyx}
\end{figure*}

{\bf{Qualitative Results.}}
We display the qualitative results of VoD and Astyx validation set in Fig.\ref{fig_qualitative} and Fig.\ref{fig_qualitative_astyx}. We picture the ground truth object on (a) images, the prediction results of (b) PointPillars, and (c) our proposed \textit{MVFAN}. The predicted 3D bounding boxes are painted in \textcolor{red}{red} only in point clouds. The ground truth objects, \textit{cars}, \textit{pedestrians} and \textit{cyclists} are colored in \textcolor{green}{green}, \textcolor{cyan}{blue} and \textcolor{yellow}{yellow} and together with images as references.

% which starkly contrast the detection results of PointPillars and our \textit{MVFAN}. Specifically, our \textit{MVFAN} produces better predictions with fewer false positives and more precise object orientation. 

\section{Conclusion}\label{sec_conclusion}
In summary, our proposed Multi-View Feature Assisted Network is a novel single-stage framework that can effectively utilize both geometric and Doppler velocity and reflectivity measured by the 4D radar sensor. Our method addresses the issue of insufficient feature utilization by employing complementary features from multi-view. We introducing a novel Position Map Generation module to reweigh foreground and background points considering the irregularity of radar point clouds. Importantly, we introduce a novel Radar Feature Assisted backbone, which is the first to fully leverage the valuable information provided by the Doppler velocity and RCS reflectivity in 4D radar data for 3D object detection while existing radar-based 3D object detection methods have largely ignored these features. Our ablation studies validate the significant impact of incorporating these features into the framework, highlighting the importance of considering this information for accurate object detection. Our extensive experiments on Astyx and VoD datasets demonstrate that our proposed method significantly outperforms state-of-the-art 3D object detection methods. 
% The majority of existing studies either are designed for detecting 2D and 3D objects using only raw radar points or attempt to incorporate other modalities for object detection but are inaccurate. In this paper, we present \textit{MVFAN}, a multi-view feature assisted network for 4D radar object detection. By meticulously designing a multi-view feature extraction module and a radar feature assisted backbone that extract metaphysical radar features, we are able to promote the detection performance and demonstrate the effectiveness on the VoD benchmark. The improvement of our model manifests the importance of learning multi-perspective point-wise representation and the analytical application of radar features.

\bibliographystyle{egbib}
\bibliography{egbib}

\begin{thebibliography}{10}
\providecommand{\url}[1]{\texttt{#1}}
\providecommand{\urlprefix}{URL }
\providecommand{\doi}[1]{https://doi.org/#1}

\bibitem{RadarRobotCarDatasetICRA2020}
Barnes, D., Gadd, M., Murcutt, P., Newman, P., Posner, I.: The oxford radar robotcar dataset: A radar extension to the oxford robotcar dataset. In: 2020 IEEE International Conference on Robotics and Automation (ICRA). pp. 6433--6438. IEEE (2020)

\bibitem{SeeThroughFog}
Bijelic, M., Gruber, T., Mannan, F., Kraus, F., Ritter, W., Dietmayer, K., Heide, F.: Seeing through fog without seeing fog: Deep multimodal sensor fusion in unseen adverse weather. In: Proceedings of the IEEE/CVF Conference on Computer Vision and Pattern Recognition. pp. 11682--11692 (2020)

\bibitem{nuScenes}
Caesar, H., Bankiti, V., Lang, A.H., Vora, S., Liong, V.E., Xu, Q., Krishnan, A., Pan, Y., Baldan, G., Beijbom, O.: nuscenes: A multimodal dataset for autonomous driving. In: Proceedings of the IEEE/CVF conference on computer vision and pattern recognition. pp. 11621--11631 (2020)

\bibitem{2DCar}
Danzer, A., Griebel, T., Bach, M., Dietmayer, K.: 2d car detection in radar data with pointnets. In: 2019 IEEE Intelligent Transportation Systems Conference (ITSC). pp. 61--66. IEEE (2019)

\bibitem{hollow_3D}
Deng, J., Zhou, W., Zhang, Y., Li, H.: From multi-view to hollow-3d: Hallucinated hollow-3d r-cnn for 3d object detection. IEEE Transactions on Circuits and Systems for Video Technology  \textbf{31}(12),  4722--4734 (2021)

\bibitem{kitti}
Geiger, A., Lenz, P., Urtasun, R.: Are we ready for autonomous driving? the kitti vision benchmark suite. In: 2012 IEEE conference on computer vision and pattern recognition. pp. 3354--3361. IEEE (2012)

\bibitem{resnet}
He, K., Zhang, X., Ren, S., Sun, J.: Deep residual learning for image recognition. In: Proceedings of the IEEE conference on computer vision and pattern recognition. pp. 770--778 (2016)

\bibitem{CramNet}
Hwang, J.J., Kretzschmar, H., Manela, J.M., Rafferty, S.M., Armstrong-Crews, N.L., Chen, T., Anguelov, D.: Cramnet: Camera-radar fusion with ray-constrained cross-attention for robust 3d object detection. In: European Conference on Computer Vision (2022)

\bibitem{T-RODNet}
Jiang, T., Zhuang, L., An, Q., Wang, J., Xiao, K., Wang, A.: T-rodnet: Transformer for vehicular millimeter-wave radar object detection. IEEE Transactions on Instrumentation and Measurement  \textbf{72},  1--12 (2022)

\bibitem{Joint_3D}
Ku, J., Mozifian, M., Lee, J., Harakeh, A., Waslander, S.L.: Joint 3d proposal generation and object detection from view aggregation. In: 2018 IEEE/RSJ International Conference on Intelligent Robots and Systems (IROS). pp.~1--8. IEEE (2018)

\bibitem{pointpillar}
Lang, A.H., Vora, S., Caesar, H., Zhou, L., Yang, J., Beijbom, O.: Pointpillars: Fast encoders for object detection from point clouds. In: Proceedings of the IEEE/CVF conference on computer vision and pattern recognition. pp. 12697--12705 (2019)

\bibitem{focalloss}
Lin, T.Y., Goyal, P., Girshick, R., He, K., Doll{\'a}r, P.: Focal loss for dense object detection. In: Proceedings of the IEEE international conference on computer vision. pp. 2980--2988 (2017)

\bibitem{SSD}
Liu, W., Anguelov, D., Erhan, D., Szegedy, C., Reed, S., Fu, C.Y., Berg, A.C.: Ssd: Single shot multibox detector. In: Computer Vision--ECCV 2016: 14th European Conference, Amsterdam, The Netherlands, October 11--14, 2016, Proceedings, Part I 14. pp. 21--37. Springer (2016)

\bibitem{3DCTN}
Lu, D., Xie, Q., Gao, K., Xu, L., Li, J.: 3dctn: 3d convolution-transformer network for point cloud classification. IEEE Transactions on Intelligent Transportation Systems  \textbf{23}(12),  24854--24865 (2022)

\bibitem{Astyx}
Meyer, M., Kuschk, G.: Automotive radar dataset for deep learning based 3d object detection. In: 2019 16th european radar conference (EuRAD). pp. 129--132. IEEE (2019)

\bibitem{3D_Radar_Camera}
Meyer, M., Kuschk, G.: Deep learning based 3d object detection for automotive radar and camera. In: 2019 16th European Radar Conference (EuRAD). pp. 133--136. IEEE (2019)

\bibitem{Graph3D}
Meyer, M., Kuschk, G., Tomforde, S.: Graph convolutional networks for 3d object detection on radar data. In: Proceedings of the IEEE/CVF International Conference on Computer Vision. pp. 3060--3069 (2021)

\bibitem{RRPN}
Nabati, R., Qi, H.: Rrpn: Radar region proposal network for object detection in autonomous vehicles. In: 2019 IEEE International Conference on Image Processing (ICIP). pp. 3093--3097. IEEE (2019)

\bibitem{CenterFusion}
Nabati, R., Qi, H.: Centerfusion: Center-based radar and camera fusion for 3d object detection. In: Proceedings of the IEEE/CVF Winter Conference on Applications of Computer Vision. pp. 1527--1536 (2021)

\bibitem{kradar}
Paek, D.H., Kong, S.H., Wijaya, K.T.: K-radar: 4d radar object detection for autonomous driving in various weather conditions. In: Thirty-sixth Conference on Neural Information Processing Systems Datasets and Benchmarks Track (2022), \url{https://openreview.net/forum?id=W_bsDmzwaZ7}

\bibitem{RoadUserDetection}
Palffy, A., Dong, J., Kooij, J.F., Gavrila, D.M.: Cnn based road user detection using the 3d radar cube. IEEE Robotics and Automation Letters  \textbf{5}(2),  1263--1270 (2020)

\bibitem{3DRadarCube}
Palffy, A., Dong, J., Kooij, J.F., Gavrila, D.M.: Cnn based road user detection using the 3d radar cube. IEEE Robotics and Automation Letters  \textbf{5}(2),  1263--1270 (2020)

\bibitem{VOD}
Palffy, A., Pool, E., Baratam, S., Kooij, J.F.P., Gavrila, D.M.: Multi-class road user detection with 3+1d radar in the view-of-delft dataset. IEEE Robotics and Automation Letters  \textbf{7}(2),  4961--4968 (2022). \doi{10.1109/LRA.2022.3147324}

\bibitem{pointnet}
Qi, C.R., Su, H., Mo, K., Guibas, L.J.: Pointnet: Deep learning on point sets for 3d classification and segmentation. In: Proceedings of the IEEE conference on computer vision and pattern recognition. pp. 652--660 (2017)

\bibitem{MVD}
Qian, K., Zhu, S., Zhang, X., Li, L.E.: Robust multimodal vehicle detection in foggy weather using complementary lidar and radar signals. In: Proceedings of the IEEE/CVF Conference on Computer Vision and Pattern Recognition. pp. 444--453 (2021)

\bibitem{Unet}
Ronneberger, O., Fischer, P., Brox, T.: U-net: Convolutional networks for biomedical image segmentation. In: Medical Image Computing and Computer-Assisted Intervention--MICCAI 2015: 18th International Conference, Munich, Germany, October 5-9, 2015, Proceedings, Part III 18. pp. 234--241. Springer (2015)

\bibitem{schumann2018semantic}
Schumann, O., Hahn, M., Dickmann, J., W{\"o}hler, C.: Semantic segmentation on radar point clouds. In: 2018 21st International Conference on Information Fusion (FUSION). pp. 2179--2186. IEEE (2018)

\bibitem{PV-RCNN}
Shi, S., Guo, C., Jiang, L., Wang, Z., Shi, J., Wang, X., Li, H.: Pv-rcnn: Point-voxel feature set abstraction for 3d object detection. In: Proceedings of the IEEE/CVF conference on computer vision and pattern recognition. pp. 10529--10538 (2020)

\bibitem{PointRCNN}
Shi, S., Wang, X., Li, H.: Pointrcnn: 3d object proposal generation and detection from point cloud. In: Proceedings of the IEEE/CVF conference on computer vision and pattern recognition. pp. 770--779 (2019)

\bibitem{AM3Net}
Wang, J., Li, J., Shi, Y., Lai, J., Tan, X.: Am$^3$net: Adaptive mutual-learning-based multimodal data fusion network. IEEE Transactions on Circuits and Systems for Video Technology  \textbf{32}(8),  5411--5426 (2022)

\bibitem{rodnet}
Wang, Y., Jiang, Z., Li, Y., Hwang, J.N., Xing, G., Liu, H.: Rodnet: A real-time radar object detection network cross-supervised by camera-radar fused object 3d localization. IEEE Journal of Selected Topics in Signal Processing  \textbf{15},  954--967 (2021)

\bibitem{Pillar-based}
Wang, Y., Fathi, A., Kundu, A., Ross, D.A., Pantofaru, C., Funkhouser, T., Solomon, J.: Pillar-based object detection for autonomous driving. In: Computer Vision--ECCV 2020: 16th European Conference, Glasgow, UK, August 23--28, 2020, Proceedings, Part XXII 16. pp. 18--34. Springer (2020)

\bibitem{mmWaveRadar}
Wei, Z., Zhang, F., Chang, S., Liu, Y., Wu, H., Feng, Z.: Mmwave radar and vision fusion for object detection in autonomous driving: A review. Sensors  \textbf{22}(7) (2022). \doi{10.3390/s22072542}, \url{https://www.mdpi.com/1424-8220/22/7/2542}

\bibitem{RPFA-Net}
Xu, B., Zhang, X., Wang, L., Hu, X., Li, Z., Pan, S., Li, J., Deng, Y.: Rpfa-net: A 4d radar pillar feature attention network for 3d object detection. In: 2021 IEEE International Intelligent Transportation Systems Conference (ITSC). pp. 3061--3066. IEEE (2021)

\bibitem{second}
Yan, Y., Mao, Y., Li, B.: Second: Sparsely embedded convolutional detection. Sensors  \textbf{18}(10), ~3337 (2018)

\bibitem{RadarNet}
Yang, B., Guo, R., Liang, M., Casas, S., Urtasun, R.: Radarnet: Exploiting radar for robust perception of dynamic objects. In: Computer Vision--ECCV 2020: 16th European Conference, Glasgow, UK, August 23--28, 2020, Proceedings, Part XVIII 16. pp. 496--512. Springer (2020)

\bibitem{TJ4DRadSet}
Zheng, L., Ma, Z., Zhu, X., Tan, B., Li, S., Long, K., Sun, W., Chen, S., Zhang, L., Wan, M., et~al.: Tj4dradset: A 4d radar dataset for autonomous driving. In: 2022 IEEE 25th International Conference on Intelligent Transportation Systems (ITSC). pp. 493--498. IEEE (2022)

\end{thebibliography}





% LNCS Homepage, \url{http://www.springer.com/lncs}. Last accessed 4
% \end{thebibliography}

\end{document}